\documentclass[journal,12pt]{IEEEtran}
\usepackage{amsmath,amsfonts}
\usepackage{algorithmic}
\usepackage{algorithm}
\usepackage{array}
\usepackage[caption=false,font=normalsize,labelfont=sf,textfont=sf]{subfig}
\usepackage{textcomp}
\usepackage{stfloats}
\usepackage{url}
\usepackage{verbatim}
\usepackage{graphicx}
\usepackage{cite}
\usepackage{multirow}
\usepackage{rotating}
\usepackage{multicol}
\usepackage{float}
\usepackage{textcomp}
\usepackage{stfloats}
\usepackage{array}
\usepackage{svg}
\usepackage{siunitx}
\usepackage{xcolor}
\usepackage{multirow}
\usepackage{graphicx}
\usepackage{color, soul, colortbl}
\usepackage{nicematrix}
\usepackage{colortbl} 
\usepackage{url}
\usepackage{hyperref}
\usepackage{mathtools}
\usepackage{epstopdf}

\DeclareUnicodeCharacter{2061}{}

\begin{document}
\title{Long-Range Vision-Based UAV-assisted Localization for Unmanned Surface Vehicles}

\author{Waseem Akram$^{1}$, Siyuan Yang$^{2}$, Hailiang Kuang$^{2}$, Xiaoyu He$^{2}$, Muhayy Ud Din$^{1}$,\\ Yihao Dong$^{2}$, Defu Lin$^{2}$, Lakmal Seneviratne$^{1}$, Shaoming He$^{2}$$^{*}$ and Irfan Hussain$^{1}$$^{*}$                                                     
\thanks{ $^{1}$Khalifa University Center for Autonomous Robotic Systems (KUCARS), Khalifa University, United Arab Emirates.}%
\thanks{$^{2}$School of Aerospace Engineering, Beijing Institute of Technology, China}
\thanks{$^{*}$ Corresponding Author, Email:shaoming.he@bit.edu.cn, irfan.hussain@ku.ac.ae}
\thanks{The first two authors have equal contribution.}
}



\maketitle

\begin{abstract}
The global positioning system (GPS) has become an indispensable
navigation method for field operations with unmanned surface
vehicles (USVs) in marine environments. However, GPS may not always be
available outdoors because it is vulnerable to natural interference and malicious jamming attacks. Thus, an alternative navigation
system is required when the use of GPS is restricted or prohibited. To this end, we present a novel method that utilizes an Unmanned Aerial Vehicle (UAV) to assist in localizing USVs in GNSS-restricted marine environments. In our approach, the UAV flies along the shoreline at a consistent altitude, continuously tracking and detecting the USV using a deep learning-based approach on camera images. Subsequently, triangulation techniques are applied to estimate the USV's position relative to the UAV, utilizing geometric information and datalink range from the UAV. We propose adjusting the UAV's camera angle based on the pixel error between the USV and the image center throughout the localization process to enhance accuracy. Additionally, visual measurements are integrated into an Extended Kalman Filter (EKF) for robust state estimation.
To validate our proposed method, we utilize a USV equipped with onboard sensors and a UAV equipped with a camera. A heterogeneous robotic interface is established to facilitate communication between the USV and UAV. We demonstrate the efficacy of our approach through a series of experiments conducted during the ``Muhammad Bin Zayed International Robotic Challenge (MBZIRC-2024)'' in real marine environments, incorporating noisy measurements and ocean disturbances. The successful outcomes indicate the potential of our method to complement GPS for USV navigation.

\end{abstract}

\begin{IEEEkeywords}
Vision-based localization, Position estimation, GNSS denial, Unmanned surface vehicles, Extended kalman filter 
\end{IEEEkeywords}

\section{Introduction}

The maritime sector is increasingly focusing on Unmanned Surface Vehicles (USVs), which possess the ability to autonomously carry out navigation missions \cite{han2019coastal}. These vehicles have gained significant attention due to their diverse and impactful applications, including the exploration of marine resources \cite{li2024multi}, oceanographic mapping \cite{zhao2024optimal,carilli2024applying}, and the inspection and monitoring of coastal and offshore structures \cite{jiao2024vision,campos2024nautilus}, ports \cite{bojke2024application}, and more \cite{volden2022vision,vasilijevic2017coordinated}. In marine dynamic environments, accurate localization of USVs (e.g., to estimate one’s position and orientation with respect to surrounding environments) is crucial \cite{usvlocrev}. The USVs require location information to make decisions related to control, navigation, collision avoidance, and path planning \cite{garofano2024obstacle}. Therefore, localization is considered a fundamental capability for USVs engaged in tracking, exploration, or monitoring of the marine environment \cite{int8,zereik2018challenges}.

\begin{figure}
\centering
\includegraphics[width=\columnwidth]{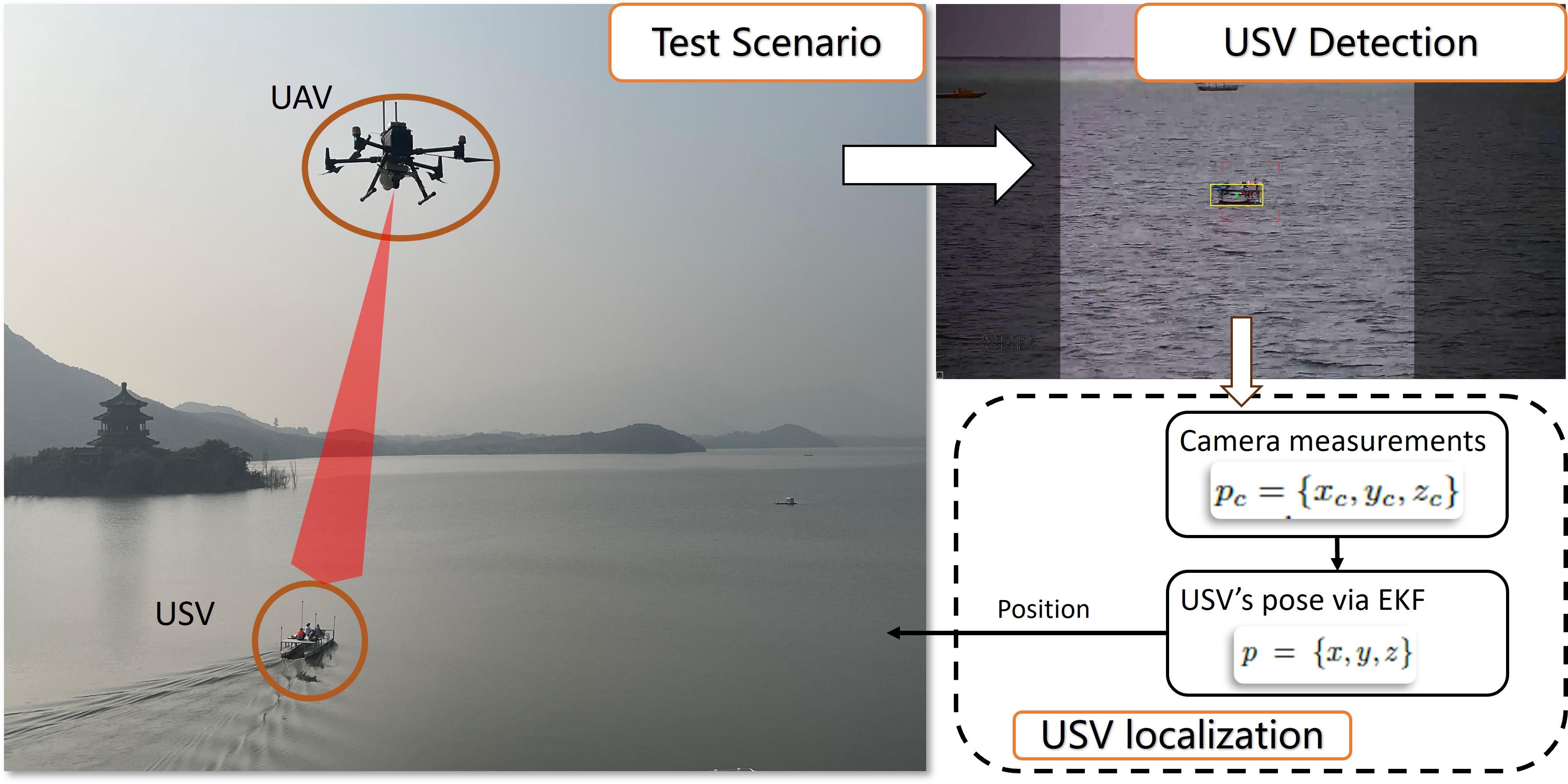}
\caption{Project concept. A heterogeneous robotic solution for USV localization in a GPS-free marine environment. The system consists of an UAV that searches and detects the USV from a fixed altitude, taking vision-based techniques to extract the USV position in camera images. Subsequently, an EKF is used estimate the final USV's positions that needed for the USV navigation and target tracking.}
\label{fig:frontimg}
\end{figure}

Due to technological advancements, an integrated navigation system, combining an Inertial Navigation System (INS) with the Global Positioning System (GPS), has been employed for navigation and localization for USVs in the marine environment \cite{roed23dem}. However, the instability of GPS signals in the dynamic marine environment, due to many reasons such as signal blockages, multipath reflection, and jamming, often leads to weak or missing signals, resulting in reduced position accuracy \cite{tabish2024maritime}. To enhance the navigation system, integrating additional sensors such as Droplet Velocity Log (DVL) and Radar for position measurements instead of relying solely on GPS becomes effective. Nonetheless, this approach inevitably increases the overall navigation cost \cite{liu23vio}.

The integration of DVL and radar systems for USV navigation offers enhanced capabilities in challenging marine environments \cite{bae2023survey}. DVL provides real-time velocity measurements by using acoustic signals aiding in precise navigation, especially in GNSS-denied environments. However, drift problems in DVL systems can arise due to cumulative errors in velocity measurements over time, leading to inaccurate position estimates \cite{jiang2024low}. Factors such as sensor misalignment, calibration errors, and variations in water properties can contribute to drift, impacting the overall navigation accuracy. On the other hand, radar systems offer the capability of obstacle detection and navigation in low visibility conditions \cite{ma2017radar}. They can identify other vessels, landmasses, or structures, enabling the USV to navigate safely. However, radars also have poor performance in close proximity and in adverse weather conditions \cite{maradar,xu2024real}.  

Recently, there has been an increased interest in developing and using vision-based localization systems because they are more robust, reliable, and cheaper than other sensor-based localization systems, e.g., acoustic or laser-based systems \cite{liu23vio,volden2022vision}. In literature, many studies focus on feature-based localization methods such as visual odometry or simultaneous localization and mapping (SLAM) systems because they are flexible and require no additional infrastructure in the environment \cite{volden2023development}. However, localization algorithms based on visual information may fail in some challenging environments, such as low-resolution features, low visibility conditions, or moving objects \cite{haldorai2024review}. In addition, sensor fusion has been used in marine environments \cite{guo2024usefulness}. The sensor fusion approach combines measurements or observations from multiple sensors such as IMU, camera, LiDAR, etc. \cite{meysam2024improved}. Fusing these diverse sensor inputs eliminates individual sensor limitations and enhances the USV's ability to navigate autonomously in dynamic and challenging marine conditions \cite{merveille2024advancements,fossen2009kalman}. 

One significant drawback of using the traditional methods for USV localization in long-range navigation and target tracking is the limited field of view and range. Onboard sensors, such as cameras, LiDAR, or radar, are constrained by the curvature of the Earth, environmental obstacles, and their inherent range limitations. This can result in blind spots, reduced accuracy, and delayed response times when tracking distant targets or navigating in a long range. The USV can drift with the ocean waves, which means it might still miss the target even if it's following the correct heading angle. As we know, UAVs offer an aerial viewpoint, providing a wide field of vision that is impossible to achieve from the surface level. This aerial advantage motivates the integration of UAVs to enhance USV localization, particularly in long-range marine environments\cite{santos2024cooperative}.

In this paper, we employ a heterogeneous robotic setup comprising both a USV and a UAV to localize the USV with assistance from a UAV using vision-based techniques. The UAV's camera identifies the USV from a fixed altitude along the shoreline, observing the USV within a broader marine setting. Subsequently, triangulation and geometric information are employed to determine the USV's position relative to the UAV using visual observations and the range of datalinks on UAV and USV as the distance between them. Moreover, we propose a method for controlling the UAV's camera orientation to center the USV within its field of view, thereby obtaining precise pose information. We integrate USV positions with EKF to enhance localization accuracy, resulting in a more robust solution based on visual measurements. A conceptual framework of the work is shown in Fig.~\ref{fig:frontimg}. This work was developed as a part of the ``Muhammad Bin Zayed International Robotics Challenge (MBZIRC-2024)'', held in Abu Dhabi, United Arab Emirates \cite{MBZIRC}. The main goal of this challenge was to develop a heterogeneous robotic solution consisting of both UAVs and USVs for tasks involving maritime monitoring and intervention in a GNSS-denied environment. The proposed method for localizing USVs was implemented and validated during the competition. Additionally, we illustrated how camera measurements can complement localization solutions in a long-range marine environment. Consequently, our work contributes to advancing and validating GNSS-independent localization systems for USVs in marine operations. 


The key contributions of this work are as follows:
\begin{itemize}
    \item To overcome the limitations of sometimes not having a GPS signal, we propose a heterogeneous framework consisting of a USV and a UAV for position estimation of the USV. 
    \item Integration of multiple sensors, including cameras and datalink mounted on UAV for precise positioning and orientation determination of USV employing vision-based localization in order to reduce the limitation of onboard sensors-based localization.
     \item Demonstration of the effectiveness of the innovative approach, combining UAV and USV technologies, in accurately localizing USV even in GNSS-denied marine environments.
    \item Conducting a series of experiments to validate the proposed method for USV localization in real marine experiments.

\end{itemize}

\section{Related Work}
Currently, a range of methodologies and technologies employed in USV localization, emphasizing sensor fusion \cite{yeong2021sensor,gupta2022simultaneous}, vision-based techniques \cite{fayyad2020deep}, radar integration \cite{shan2021lidar}, and deep learning \cite{choi2021robust,xu2017deep} applications to enhance accuracy, robustness, and reliability in challenging maritime environments. Each approach contributes unique insights and advancements towards achieving effective and efficient USV localization capabilities. However, these studies exhibit notable limitations in the context of USV localization. Firstly, they are constrained by range, operating within limited geographical areas. Many of these methods rely on vision-based techniques using USV onboard camera imagery, which becomes impractical when the USV must navigate the broader region.
Additionally, achieving feature-based localization becomes challenging due to poor visibility, low light conditions, and inadequate texture information in imaging data. Hence, we propose a UAV-assisted vision-based localization for USVs to enhance navigation capabilities in GPS-restricted environments. In contrast to existing approaches, our work aims to address these challenges by integrating UAV and USV equipped with diverse sensors for localization, a combination rarely experimentally demonstrated in real-world marine settings under GNSS-denied environments.
In the following, we briefly review different approaches, such as Radar-based, LiDAR-based, and Vision-based, for USV localization in GNSS-denied environments.

\subsection{Radar-based Approaches}

Radar-based localization techniques for USVs have emerged as practical solutions, particularly in GPS-denied or challenging maritime environments. Radar systems offer unique capabilities for detecting and mapping surrounding obstacles and features, which can be used for accurate USV positioning and navigation. Several notable studies have explored radar-based USV localization methods. For example, Han et al. \cite{han2019coastal} proposed a radar-centric approach for USV localization and navigation, adopting the SLAM paradigm. Their method involves extracting coastline contours from radar-acquired images using image processing techniques. These extracted contours serve as landmark features for localization, and an EKF-based algorithm is employed for estimating vehicle positions based on these features. The experimental results highlight this approach's computational efficiency and effectiveness compared to traditional point-cloud methods. Ma et al. \cite{maradar} introduced a technique that leverages the fusion of radar and satellite imagery for USV localization. Their approach uses computer vision methodologies to extract coastline features from radar and satellite images. By developing an image registration technique that accounts for horizontal and vertical perspectives captured in the input images, they demonstrated the viability of this approach with an average error of 9.77 meters in USV positioning. Dagdilelis et al. \cite{DAGDILELIS2022243} presented a novel radar-based method for localizing USVs using sea chart information. They proposed utilizing radar detection to identify underwater buoys and matching these buoys with entries from electronic navigation charts. Subsequently, triangulation and trilateration methods are applied for precise pose estimation. The simulation results in the Great Belt region in Denmark showed promising reductions in uncertainty regarding pose and heading estimation. However, further research in this area may focus on improving radar sensor technologies, optimizing algorithms for real-time processing, and integrating radar systems with other sensor modalities to enhance further USV localization performance and reliability.

\subsection{LiDAR-based Approaches}

LiDAR-based localization methods for USVs offer promising solutions for navigating in GPS-denied environments by utilizing laser scanning technology to generate high-resolution 3D maps of the surroundings. Several studies have explored LiDAR-based USV localization techniques, demonstrating their effectiveness and applicability in challenging maritime scenarios. Shen et al. \cite{shen2023lidar} introduced a novel approach for USV localization by integrating LiDAR SLAM with GNSS/INS systems. Their method employs a dynamic switching strategy to transition to LiDAR SLAM positioning when GPS signals are unavailable or unreliable. Position and heading estimates are refined using the EKF algorithm. Experimental results showed a significant reduction (55.4\%) in position error compared to traditional Kalman filter algorithms, highlighting the potential of LiDAR-based localization for USVs.
The study in \cite{ZulkifliMd} discussed an autonomous SLAM navigation, path planning, and collision avoidance system for the USV, equipped with a Velodyne 3D VLP16 lidar sensor and Axis PTZ camera. Using the Robot Operating System (ROS) navigation stack, the USV demonstrates successful autonomous navigation, path planning, and obstacle avoidance in marine environments, generating detailed maps for pipeline inspection. Similarly, the studies in \cite{skjellaug2020feature,lu2021real,skjellaug2020feature} also proposed using LiDAR technology for the USV localization in marine environments. Although, LiDAR-based USV localization methods offer advantages such as high accuracy, independence from external signals (like GPS), and suitability for mapping complex environments with obstacles. However, many challenges exist, including the cost and complexity of LiDAR systems and the need for robust algorithms to handle real-time processing of dense point cloud data in dynamic maritime environments.

\subsection{Vision-based Approaches}

Over the past few years, significant efforts have been made regarding USV localization and navigation in marine complex environments. One approach that has gained popularity in this context is the implementation of vision-based localization. For instance, Liu et al. \cite{liu23vio} proposed a visual-inertial odometry (VIO) technique for USV localization in GPS-restricted environments. They utilized cameras to capture point and line features along bridge walls, integrating this visual data with inertial measurements for real-time position estimation. While effective, the method's computational complexity is a noted challenge. Roedele et al. \cite{roed23dem} introduced a monocular camera imaging method within USV operating zones. By matching camera features with synthetic images from a digital elevation model (DEM), they derived 3-dimensional position estimates. However, practical deployment feasibility remains a concern. Volden et al. \cite{volden2022vision} explored a stereo-vision approach for USV localization during docking maneuvers. They utilized stereo cameras to detect and triangulate ArUco tags at docking stations, integrating deep learning and feature-matching techniques. While practical, further testing across diverse weather conditions is required for robustness validation. Hu et al. \cite{hu2023semantic} leveraged lidar semantic and geometric data alongside deep learning models for USV localization during docking and departing scenarios. Their approach identified and mitigated the influence of dynamic objects on localization accuracy, achieving superior performance compared to traditional GPS systems.

\begin{figure*}[t]
\centering
\includegraphics[width=1\linewidth]{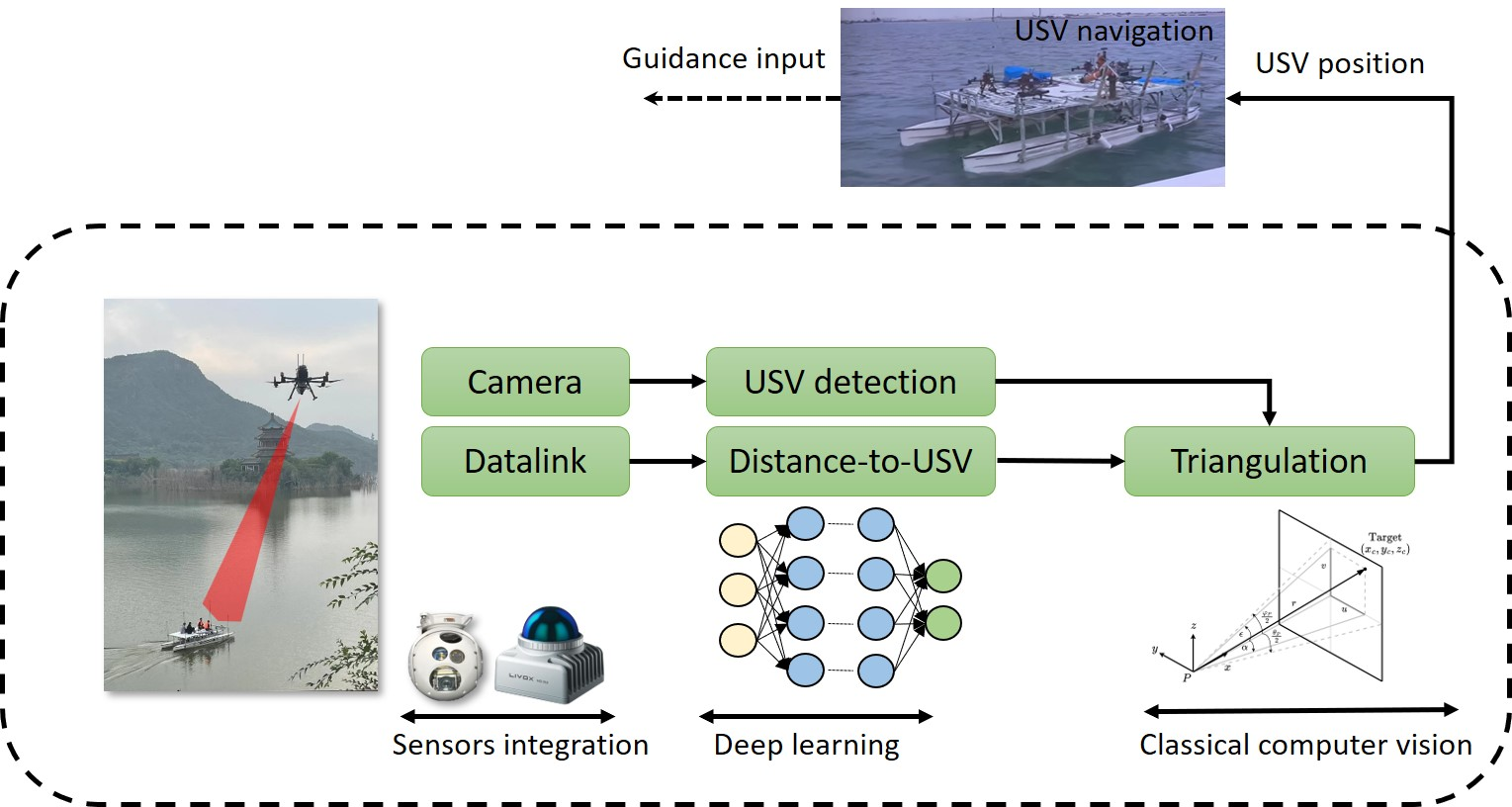}
\caption{Proposed block diagram of the system. The UAV searches and detects the USV from a fixed altitude. The camera and datalink input are used for USV position estimation, that are required for USV navigation in the marine environment.
}
\label{fig:method}
\end{figure*}

\section{Proposed Approach} \label{sec:proposed}


\subsection{Solution overview}
The proposed framework for USV localization consists of two main components: the UAV and the USV. The UAV is equipped with a high-definition movable camera, which it uses to scan the open sea environment and identify target locations, such as vessels. Once a target is identified, the UAV records its position information, establishing it as the reference point. The UAV then maintains a fixed altitude and continuously scans the USV, accurately estimating its relative position. This is achieved using a state-of-the-art deep learning algorithm that enables the UAV to track and locate the USV. An Extended Kalman Filter (EKF) is employed to determine the USV's position relative to the UAV's location and altitude based on detection and geometric measurements integrated with the radio range (datalink range) measurements. With the target position provided by the UAV, the algorithm generates the USV's state matrix in the NED (North-East-Down) frame using the EKF. This data is subsequently used by conventional closed-loop control to guide the USV toward the target. An overview of the framework is shown in Fig.~\ref{fig:method}. The proposed scheme offers several advantages, notably robust USV localization in open sea environments despite external disturbances such as ocean waves and wind. Furthermore, sensor fusion techniques that integrates camera and datalink measurements, ensure precise localization in GNSS-denied environments. Next, we explore the details of the proposed method, which is divided into USV detection, USV localization, and USV positioning using EKF.

\subsection{USV detection}
\begin{figure}[t]
\centering
	\includegraphics[width=\columnwidth]{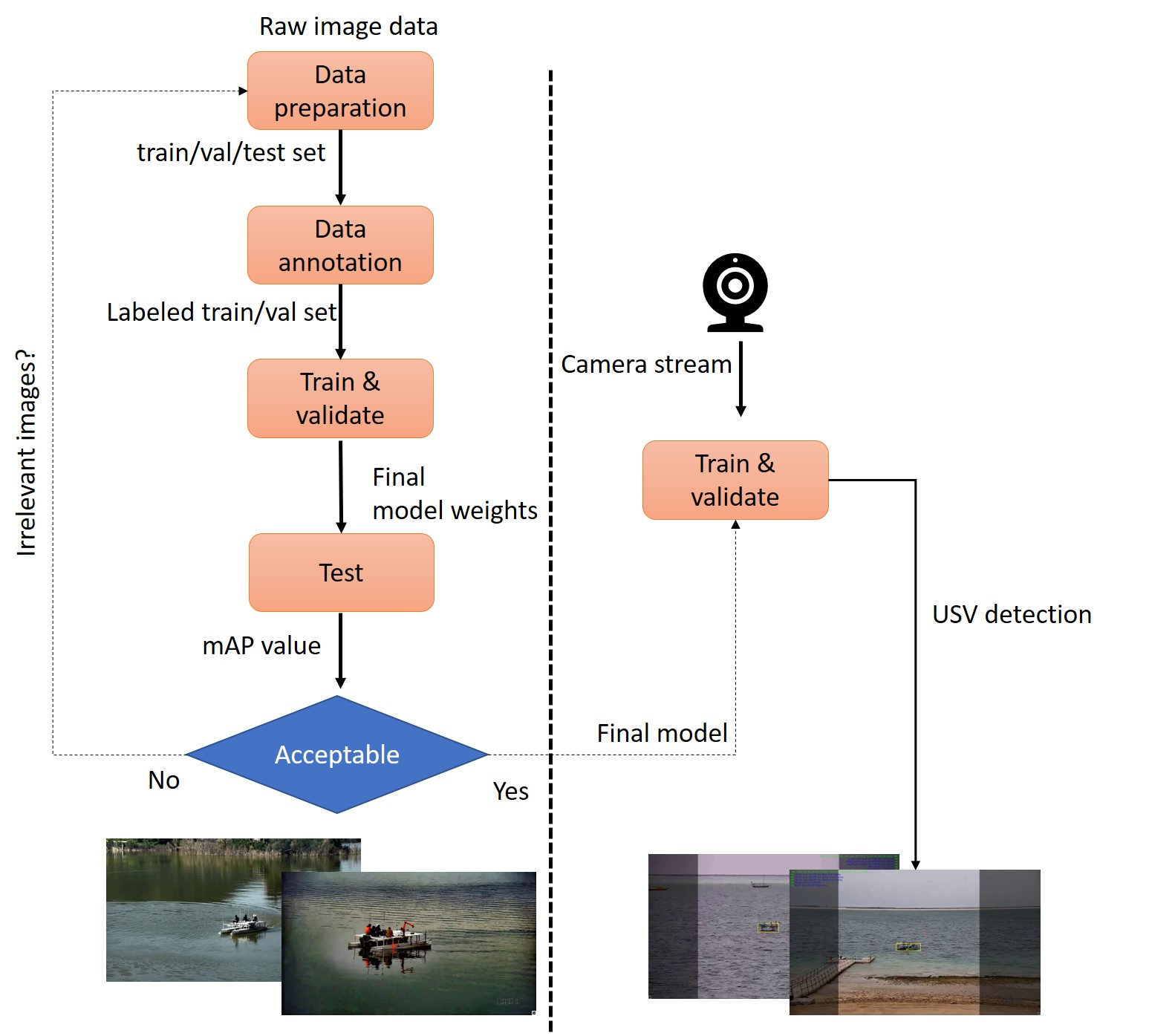}
	\caption{Steps to obtain a fined-tuned  USV detection model in real marine environment.}
\label{fig:usvtrain}
\end{figure}

This section discusses the essential steps in obtaining the USV detection model using data-driven methods. Fig. \ref{fig:usvtrain} gives an overview of the process. This illustrates the essential steps involved in refining an object recognition model. Initially, images are prepared and annotated to establish ground truth for the supervised Convolutional Neural Network (CNN) learning process. The annotated data is then introduced into the data-driven detection model, and pre-trained model weights are used to refine the model. A validation set is employed for model selection, determining the appropriate stopping point for training. Subsequently, the refined model tests unseen data to assess its accuracy, often measured by mean average precision (mAP) metrics. The final model weights are deployed for object recognition tasks during the prediction phase if the outcomes are satisfactory. Next, we discuss the preparation of the USV dataset, model training, and model testing. 

\subsubsection{Step 1- Data preparation}

\begin{figure}[t]
\centering
	\includegraphics[width=\columnwidth]{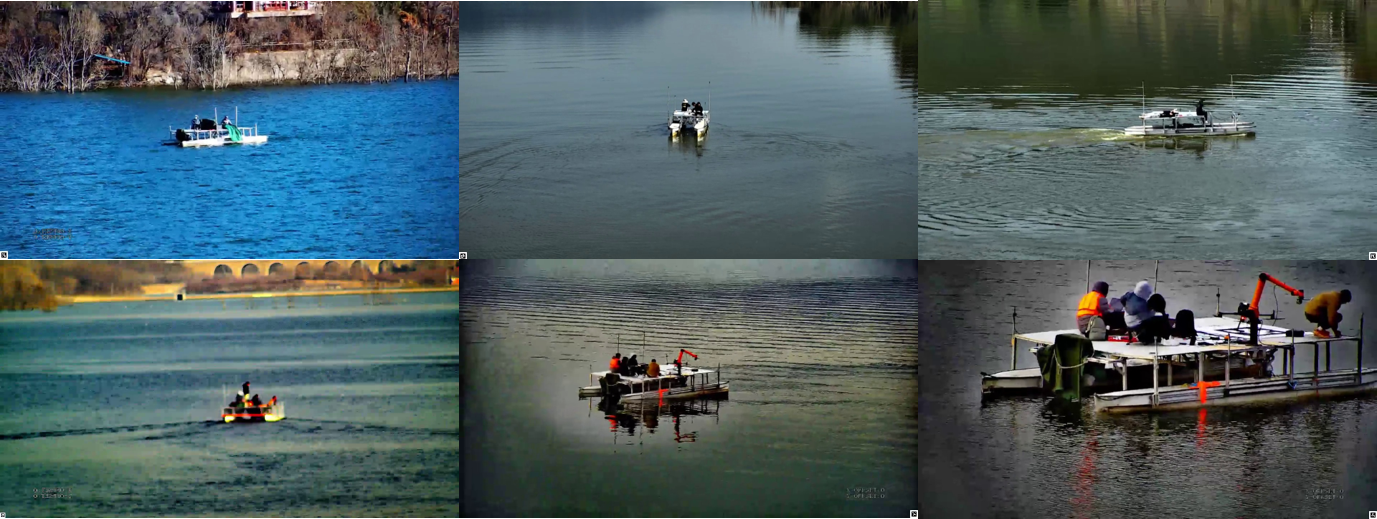}
	\caption{Dataset for USV detection collected from the UAV camera. The dataset contains USV images from different angles and distances.}
\label{fig:usvdata}
\end{figure}

The dataset plays a pivotal role in training CNN models in supervised learning. This data serves as the ground truth for the specific class the model aims to learn. During this phase, a custom dataset is collected, comprising examples of the class and their associated features, which the model is expected to learn. In our study, we conducted data collection to gather imaging data of USV from various perspectives and angles within a marine environment. A comprehensive custom dataset containing images of USVs captured in real marine settings was curated. This dataset encompasses color image data captured by a UAV while operating over the sea surface, as depicted in Fig.~\ref{fig:usvdata}. Subsequently, the dataset was randomized and divided, allocating 80\% for training purposes and 10\% each for validation and testing, respectively.

Ground truth labels guide the supervised model towards the correct answer. We use the annotation program Yolo Mark to create ground truth labels. That is, rectangle-shaped bounding boxes are dragged around the USV in the scene. Consequently, the features that help to fine-tune the model are those that recognize the USV. Notice that precise labeling is essential for the learning process. Unexpected learning often a result of inaccurate labels, e.g., only label parts of the object can be dangerous as the model interprets this as the complete object.
\subsubsection{Step 2- Data training}

For USV detection, we employ a transfer learning approach, where a pre-trained model serves as the initial point for fine-tuning toward the final detection task. The pre-trained model and its parameters are trained on the ImageNet dataset \cite{deng2009imagenet}, which comprises 15 million annotated images. We utilize pre-trained weights along with our custom dataset for model training. The original YOLOv5s and YOLOv5s6 models are employed for this custom training. YOLOv5, developed by Ultralytics \cite{yolov5}, is a deep learning model designed explicitly for object detection tasks. Its architecture primarily consists of three components: a Backbone, which forms the main network body, utilizing the CSP-Darknet53 structure in YOLOv5's design; a Neck part, connecting the backbone and the model's head, incorporating SPPF and New CSP-PAN structures in YOLOv5; and a Head part, responsible for generating the final output, utilizing the YOLOv3 Head in YOLOv5. YOLOv5 represents a significant advancement in real-time object detection models, surpassing previous iterations of the YOLO family in performance and efficiency by integrating various new features, enhancements, and training strategies. It achieved an mAP value of 72.7\% when evaluated on the COCO dataset with a 0.5 Intersection over the Union (IoU) threshold.

The hardware environment for training is characterized by Intel Core i7-12700KF CPU 3.6GHz with a single GPU of NVIDIA RTX 3090 24G memory, while Python3.8.10 and Pytorch1.8.0 configure the software environment. We trained two models for USV detection, named yolov5s6 and yolov5s. We used pre-trained models on the COCO dataset and fine-tuned them on a self-made USV dataset. For the yolov5s6 model, the batch size is 32, the image size is 1280, and the learning rate is 0.01. For the yolov5s model, the batch size is 64, the image size is 640, and the learning rate is 0.01. All models are trained with 32 epochs. 

\subsubsection{Step 3- Data testing and prediction}

\begin{figure}[t]
\centering
	\includegraphics[width=\columnwidth]{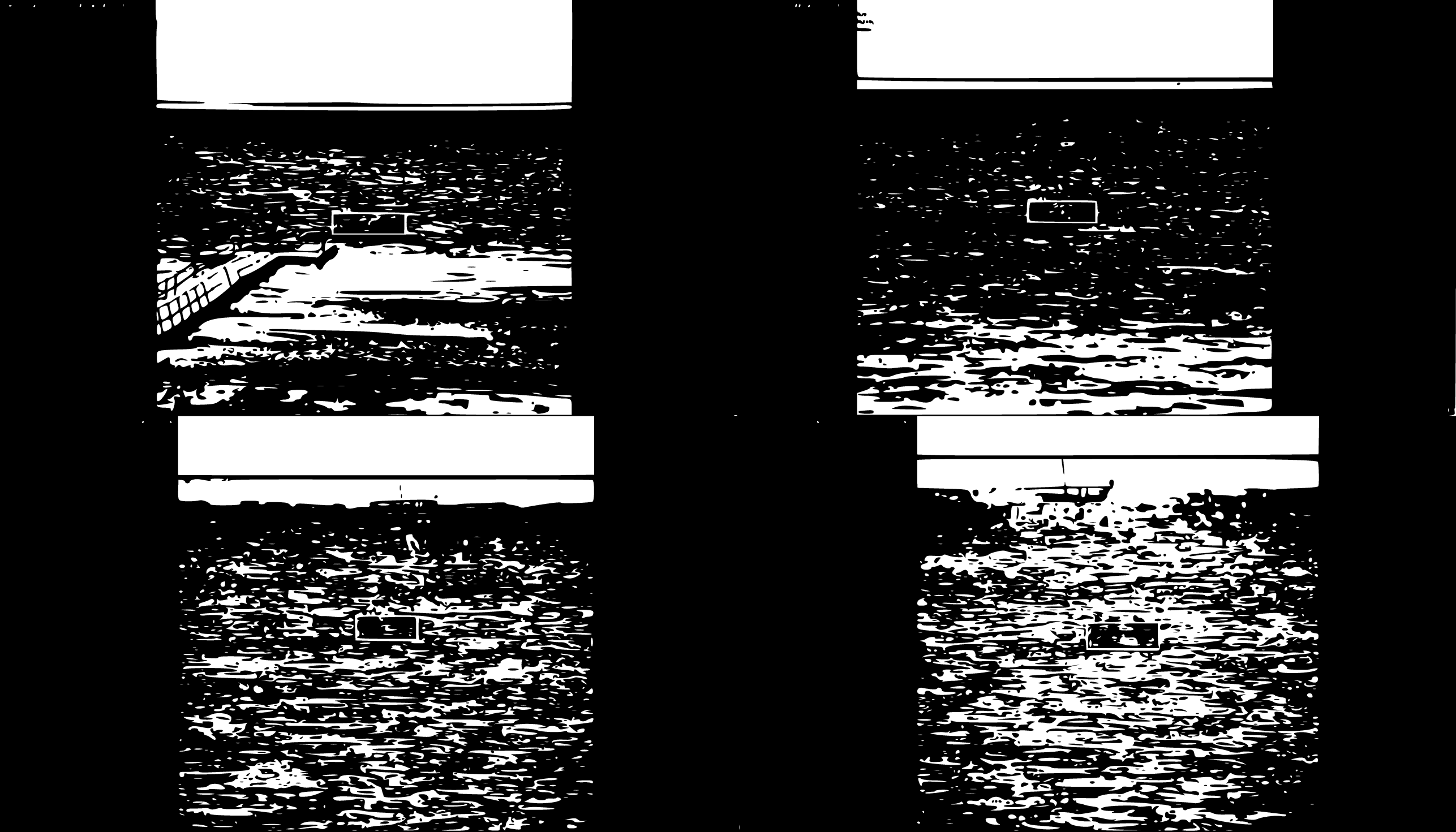}
	\caption{USV detection using YoloV5 model on the UAV's camera images during the final competition MBZIRC-2024 challenge.}
\label{fig:usvdetect}
\end{figure}

Following model training, the trained model undergoes testing on previously unseen data, referred to as the test data, to evaluate its performance. We utilize mean Average Precision (mAP) as a metric to check the performance of the trained model. Typically, a higher mAP signifies superior model performance, indicating its suitability for the tasks it was trained on. In our scenario, as the model was specifically trained for USV detection, we expect it to produce a higher mAP value during the testing phase. Upon testing the trained model on a test set extracted from the custom dataset, it achieved a mAP score of 99.05\%. Given these satisfactory results, the final trained model is deployed for USV detection in real-world experiments. During the prediction phase, we use a UAV's camera to scan the environment continuously, stream real-time data, and employ the trained model to identify USVs. A few example of the USV detection is given in Fig.~\ref{fig:usvdetect}.

\subsection{USV localization}

In this section, we present the approach used for localizing USVs. This localization process is executed by a heterogeneous robotic system consisting of both UAV and USV in a marine environment where GNSS signals are restricted. The UAV acquires visual data using its camera from a fixed altitude near the coastal area, scanning the surroundings. Initially, the UAV employs a search method to locate the USV within its field of view as shown in Fig.~\ref{fig:uavcam}. This entails controlling the camera's angle and position to scan the area effectively. Once the USV is detected within the camera's field of view, the detector identifies and extracts the bounding box of the USV object within the image frame. The results of this bounding box and camera information are used for USV pose estimation to allow smooth USV navigation along the predefined target positions. 


Vision-based pose estimation relies on the orientation and position of the camera. The UAV's camera needs to maintain the detected USV centrally within the image frame. This task is accomplished through the process involves calculating the pixel error and sending control input to the UAV's camera, allowing it to make adjustments to minimize the pixel error between the image center and the bounding box showing the detected USV. 


\subsubsection{Coordinate transformations}
In the context of USV localization and navigation, it's common to represent the USV position in the NED (North-East-Down) inertial frame. In the NED frame, the x-axis points north, the y-axis points east, and the z-axis points downward towards the Earth's center. This is a local coordinate system and is often preferred in marine navigation \cite{fossen2011handbook}. As such, we process the transformation of the camera measurement to the USV inertial frame. 

This coordinate transformation aims to estimate the USV's current position and heading angle, denoted as $p=\{x,y,z, yaw\}$, within the inertial frame as the USV navigates toward the predefined target. Estimating the USV pose involves integrating data collected from various reference frames. The following different frames are used in the coordinate transformations. 

As depicted in Table \ref{tab:frm}, the ``Inertial'' denotes the East-North-Up (ENU) frame, with the UAV takeoff position on the ground serving as the origin. In the ENU frame, East is represented by the $x$ axis, North by the $y$ axis, and Up by the $z$ axis. The ``Body'' means the Front-Left-Up (FLU) frame, originating at the UAV camera. In the FLU frame, the front is denoted by the $x$ axis, the left side by the $y$ axis, and the upward direction by the $z$ axis. The ``Pod/Camera'' illustrates the Front-Left-Up (FLU) frame, centered at the origin of the UAV's camera. In this FLU frame, the front of the camera corresponds to the $x$ axis, the left side corresponds to the $y$ axis, and the upward direction corresponds to the $z$ axis.

\begin{table}[h]
\centering
\caption{Coordinate frames used in USV pose estimation.}
\begin{tabular}{lll}
\hline
Notation & Frame  & Description  \\\hline
$I$     & Inertial &    ENU frame  \\
$B$   & Body &  FLU frame \\
$P$    & Pod/Camera &   FLU frame of camera  \\\hline

\end{tabular}\label{tab:frm}
\end{table}

\begin{figure}[t]
\centering
\includegraphics[width=1\linewidth]{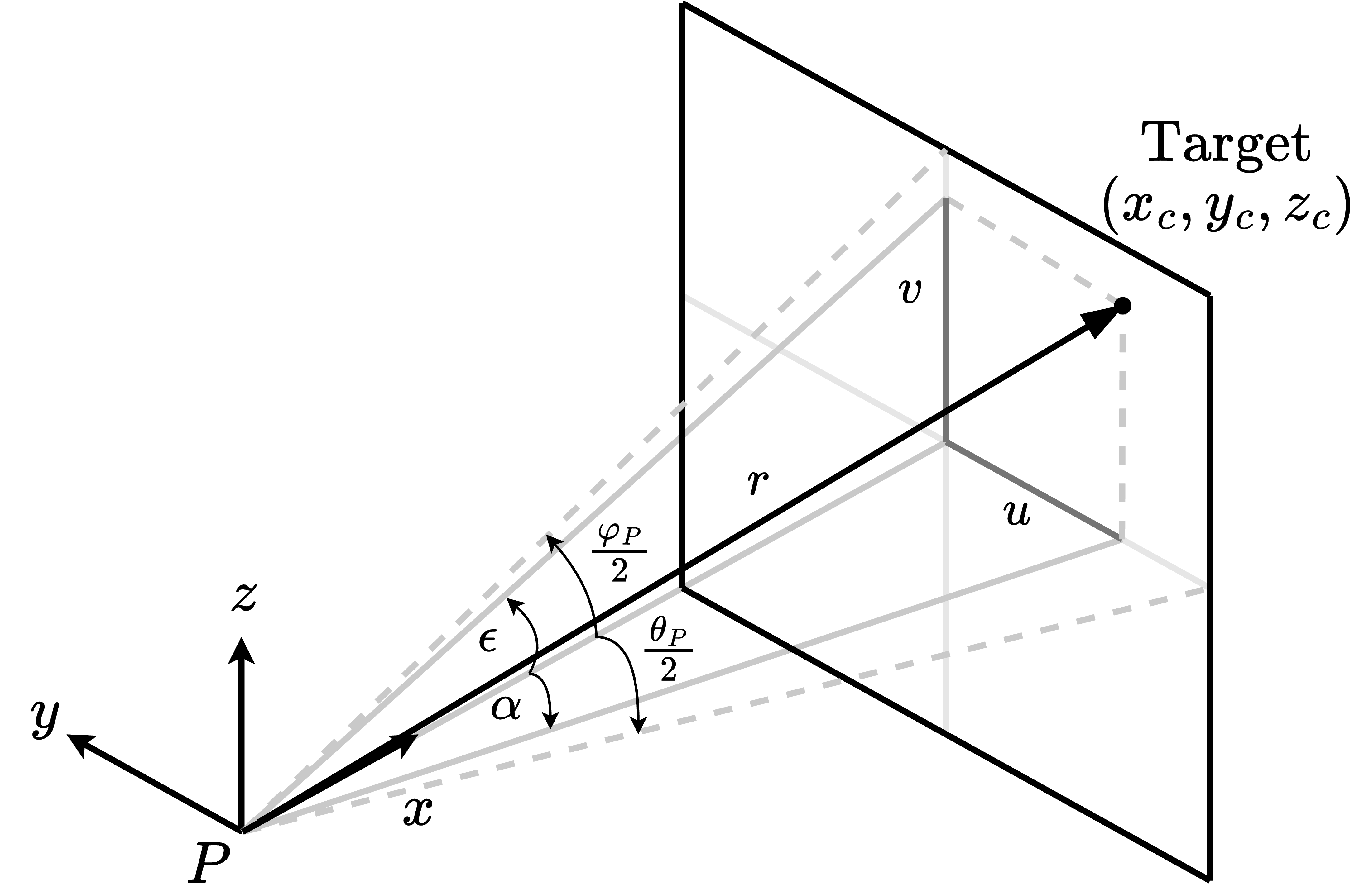}
\caption{The geometric relation of the target positions in image with respect to the camera frame. 
}
\label{fig:tri}
\end{figure}

Let \hbox{$\{r_I$, $p_I$, $y_I\}$} represent the Euler angles denoting roll, pitch, and yaw, respectively, in the \textit{I} frame of the UAV. Moreover, the camera's Euler angles in the \textit{B} frame, denoted by $\{r_B$, $p_B$, $y_B\}$, indicate the camera's roll, pitch, and yaw angles. The camera's field of view, both horizontally and vertically, is expressed as $\theta_P$ and $\varphi_P$, respectively. Additionally, we use pixel error between the detected position of the USV in the image, considering both horizontal and vertical directions relative to the center of the image. For this purpose, normalized pixel errors for horizontal and vertical directions are represented as ($u, v$)$\in [-1,1]$. Moreover, the datalink range is denoted by $r$, indicating the range between the UAV and the USV. Lastly, the UAV's position is denoted by $p_\text{uav} = [x_\text{uav}, y_\text{uav}, z_\text{uav}]^T = [x_\text{uav}, y_\text{uav}, h]^T$, where the $x_\text{uav}$ and $y_\text{uav}$ positions are assumed to be $0$, and $h$ represents a constant fixed position. The geometric relation of the target positions with respect to the camera is depicted in Fig. \ref{fig:tri}.

Our goal is to find the current position of the USV in the $I$ frame by processing each image frame. The USV target pose denoted by $p=\{x,y,z\}$ is derived by using the following steps. 
\begin{itemize}
    \item Compute Pose in camera frame $p_c=\{x_c,y_c,z_c\}$ using measurements from $\theta_p$, $\varphi_p$, $u$, $v$, and $r$
    \item Compute Pose in body frame $p_b$ using measurements from $p_c$ and $r$,
    \item Compute Pose in inertial frame $p=\{x,y,z\}$ using measurements from $p_b$, $r$, and $p_{uav}$
\end{itemize}

Initially, to obtain $p_c$, we use the object detector to identify the USV and obtain its $x$ and $y$ pixel coordinates in the image frame. Subsequently, the normalized pixel error $(u,v)$ is derived with respect to the image center. According to Fig. \ref{fig:tri}, the correlation between the Azimuth angle ($\alpha$) of the target in the $P$ frame and $u$, and the Elevation angle ($\epsilon$) of the target in the $P$ frame and $v$, is written as follows.

\begin{equation}
\begin{cases}
	\dfrac{1}{\tan \dfrac{\varphi_P}{2}} = \dfrac{v}{\tan (-\epsilon)}\\
	\dfrac{1}{\tan \dfrac{\theta_P}{2}} = \dfrac{u}{\tan (-\alpha)}.
\end{cases}
\end{equation}

Therefore, $\epsilon$ and $\alpha$ are computed as follows:

\begin{equation}\label{eq:alpha_epsi}
\begin{cases}
	\epsilon = -\arctan(v \tan\dfrac{\varphi_P}{2})\\
	\alpha = -\arctan(u \tan\dfrac{\theta_P}{2}).
\end{cases}
\end{equation}

By using $\alpha$ and $\epsilon$ measurements, the target pose $p_c$ is computed as follows:

\begin{equation}\label{eq:pc}
p_c = \begin{bmatrix}
	x_c \\ y_c \\ z_c
\end{bmatrix}
=
\begin{bmatrix}
	1 \\ \tan\alpha \\ \tan\epsilon
\end{bmatrix}
r / \sqrt{1 + \tan^2\alpha + \tan^2\epsilon}.
\end{equation}

Equation (\ref{eq:pc}) computes the position $p_c$ of the target object in the camera. Subsequently, we must process the transformation from the camera frame's $p_c$ to the body frame's $p_b$. To achieve this, let $R_B^P$ represent the rotation matrix between the camera and body frames, where $r_p$, $p_p$, and $y_p$ denote the camera's Euler angles, corresponding to roll, pitch, and yaw, respectively. Additionally, the rotation matrix between the body and Inertial frames is denoted as $R_I^B$, with $r_b$, $p_b$, and $y_b$ representing the UAV's Euler angles in the body frame. Mathematically, the relation between $p_c$ and $p_b$ is given as follows:

\begin{equation}
	p_c = R^P_B p_b
\end{equation}

Similarly, the relation between $p_b$ and $p$ is given as follows:

\begin{equation}
	p_B = R^B_I (p - p_\text{UAV}).
\end{equation}

Finally, we can obtain $p$ as follows:

\begin{equation}
\begin{aligned}
	p & = p_\text{UAV} + {R^B_I}^{-1} p_B \\
    & = p_\text{UAV} + R^I_B p_B \\
    & = p_\text{UAV} + R^I_B {R^P_B}^{-1} p_c\\
    & = p_\text{UAV} + R^I_B R^B_P p_c
\end{aligned}
\end{equation}

\subsection{USV positioning via EKF}

We base our measurements solely on geometric principles when we obtain the state variables through basic geometric relationships. In this approach, each frame of measurement provides outputs that are independent of each other, lacking any intrinsic relationship. Consequently, this independence between measurement frames leads to significant output variations. The lack of correlation between measurements from different frames can introduce uncertainty and inconsistency in the estimation process \cite{simon2006optimal}.

We utilize the EKF to address this issue. The EKF is a powerful tool in estimation theory that enables the estimation of state variables in systems with nonlinear dynamics \cite{ribeiro2004kalman}. Unlike traditional linear estimation techniques, the EKF can effectively handle nonlinear relationships between variables, making it particularly suitable for scenarios where geometric relationships alone may not suffice to capture the complexity of the system dynamics \cite{barfoot2024state}.

By employing the EKF, we aim to integrate the information from multiple measurement frames and exploit the correlations between them to improve the accuracy and reliability of our state variable estimates. The EKF achieves this by iteratively updating the state estimates based on the latest measurements while considering the system's nonlinear dynamics \cite{ding2024high}. Through this iterative process, the EKF refines the forecast, reducing the impact of measurement variations and enhancing the overall robustness of the estimation process \cite{haykin2004kalman}.

The most critical part of using EKF is model creation, based upon estimation theory principles, to derive a nonlinear transition function that accommodates unknown variables within each estimation state \cite{simon2006optimal}. Within the framework of EKF, there are two general essential models: the state model and the measurement model, expressed as follows:

\begin{equation}
\begin{aligned}
\text{State model:} \,\,\, x_{k+1}=f(X_k,u_k+w_k)\\
\text{Measurement model:} \,\,\, z_k=h(X_k+v_k)
\end{aligned}
\end{equation}

\noindent here, $x$ represents the state model comprising parameters utilized for state estimation, where $x_{k+1}$ denotes the predicted subsequent state of the model. The term $u_k$ signifies the control input, while $w_k$ pertains to noise inherent in the system. In the measurement model, $z_k$ encapsulates data from various sensors, while $v_k$ denotes Gaussian white noise.

The state variable is set as position of target in $I$ frame,  a $3\times1$ vector,
\begin{equation}
    x = [x, y, z]^T
\end{equation}

For the process model, we assume it as stationary target, i.e.,

\begin{equation}
    F = \begin{bmatrix} 1 & 0 & 0 \\ 0 & 1 & 0 \\ 0 & 0 & 1\end{bmatrix}
\end{equation}

Measurements are range $r$, camera azimuth $\alpha$, camera elevation $\epsilon$, and height of UAV $h$, i.e.,
\begin{equation}
    h(\bar {\mathbf x}) = \mathbf z = [r, \alpha, \varepsilon, h]^T
\end{equation}

Measurement equation is
\begin{equation}
\begin{aligned}
\mathbf z = \begin{bmatrix}
r \\ \alpha \\ \epsilon \\ h
\end{bmatrix}= \begin{bmatrix} 
\sqrt{x_c^2+y_c^2+z_c^2}\\ \arctan(y_c/x_c) \\ \arctan(z_c/x_c) 
\\
z + z_\text{uav}
\end{bmatrix}
\end{aligned}
\end{equation}

One key aspect of EKF is the use of the Jacobian matrix, which is essentially the first-order derivative, to linearize the non-linear functions $f$ and $h$. Jacobi matrix is:

\begin{equation}
\begin{aligned}
	\mathbf{H} = \frac{\partial h (\bar{\mathbf{x}})}{\partial (\bar{\mathbf{x}})} \Bigg |_{\bar{\mathbf{x}}}
	=
\begin{bmatrix}
\dfrac{\partial r}{\partial x} &
\dfrac{\partial r}{\partial y} &
\dfrac{\partial r}{\partial z} \\
\dfrac{\partial \alpha}{\partial x} &
\dfrac{\partial \alpha}{\partial y} &
\dfrac{\partial \alpha}{\partial z} \\
\dfrac{\partial \varepsilon}{\partial x} &
\dfrac{\partial \varepsilon}{\partial y} &
\dfrac{\partial \varepsilon}{\partial z} \\
\dfrac{\partial h}{\partial x} &
\dfrac{\partial h}{\partial y} &
\dfrac{\partial h}{\partial z} \\
\end{bmatrix}
\end{aligned}
\end{equation}

\begin{equation}
\begin{aligned}
	\mathbf{H} = \frac{\partial h (\bar{\mathbf{x}})}{\partial (\bar{\mathbf{x}})} \Bigg |_{\bar{\mathbf{x}}}
	=
\begin{bmatrix}
\dfrac{\partial r}{\partial x} &
\dfrac{\partial r}{\partial y} &
\dfrac{\partial r}{\partial z} \\
\dfrac{\partial \alpha}{\partial x} &
\dfrac{\partial \alpha}{\partial y} &
\dfrac{\partial \alpha}{\partial z} \\
\dfrac{\partial \varepsilon}{\partial x} &
\dfrac{\partial \varepsilon}{\partial y} &
\dfrac{\partial \varepsilon}{\partial z} \\
0 &
0 &
1 \\
\end{bmatrix}
\end{aligned}
\end{equation}

\begin{equation}
\begin{aligned}
\centering
\mathbf{H} =
\begin{bmatrix}
\dfrac{\partial r}{\partial p_c}\dfrac{\partial p_c}{\partial x} &
\dfrac{\partial r}{\partial p_c}\dfrac{\partial p_c}{\partial y} &
\dfrac{\partial r}{\partial p_c}\dfrac{\partial p_c}{\partial z} \\
\dfrac{\partial \alpha}{\partial p_c}\dfrac{\partial p_c}{\partial x} &
\dfrac{\partial \alpha}{\partial p_c}\dfrac{\partial p_c}{\partial y} &
\dfrac{\partial \alpha}{\partial p_c}\dfrac{\partial p_c}{\partial z} \\
\dfrac{\partial \varepsilon}{\partial p_c}\dfrac{\partial p_c}{\partial x} &
\dfrac{\partial \varepsilon}{\partial p_c}\dfrac{\partial p_c}{\partial y} &
\dfrac{\partial \varepsilon}{\partial p_c}\dfrac{\partial p_c}{\partial z} \\
0 &
0 &
1 \\
\end{bmatrix}.
\end{aligned}
\end{equation}

The first three rows of $\mathbf H$ is
\begin{equation}
\begin{aligned}
\mathbf H[0:3, 0:3]
&= 
\begin{bmatrix} 
\dfrac{\partial r}{\partial p_c} \\ 
\dfrac{\partial \alpha}{\partial p_c} \\
\dfrac{\partial \varepsilon}{\partial p_c} \\
\end{bmatrix}
\begin{bmatrix}
\dfrac{\partial p_c}{\partial x} &
\dfrac{\partial p_c}{\partial y} &
\dfrac{\partial p_c}{\partial z}
\end{bmatrix}
\end{aligned} 
\end{equation}

\begin{equation}
\begin{aligned}
&=
\begin{bmatrix}
\dfrac{x_c}r & \dfrac{y_c}r & \dfrac{z_c}r \\
\dfrac{-y_c}{x_c^2+y_c^2} & \dfrac{x_c}{x_c^2+y_c^2} & 0 \\
\dfrac{-z_c}{x_c^2+y_c^2} & 0 & \dfrac{x_c}{x_c^2+y_c^2} \\
\end{bmatrix}
R^P_B R^B_I
\\
&\coloneqq R^C_P R^P_B R^B_I.
\end{aligned} 
\end{equation}

Then we can concatenate with the last row as

\begin{equation}
\begin{aligned}
\mathbf H = \frac{\partial \mathbf z}{\partial \mathbf x}
&=
\begin{bmatrix}
&
&
\\
&
R^C_P R^P_B R^B_I&
\\
&
&
\\
0 &
0 &
1\\
\end{bmatrix}.
\end{aligned}
\end{equation}

A typical EKF model with linear $\mathbf F$ and nolinear $h(\mathbf x)$ is:

\begin{equation}
\begin{aligned}
\left.
\begin{matrix}
\bar {\mathbf x} = \mathbf F \mathbf x \\
\bar {\mathbf P} = \mathbf F \mathbf P \mathbf F^T + \mathbf Q
\end{matrix}
\quad
\right\}& \text{Prediction},
\\
\left.
\begin{matrix}
\mathbf y = \mathbf z - h (\bar {\mathbf x}) \\
\mathbf K = \bar {\mathbf P} \mathbf H^T (\mathbf H \bar{\mathbf P}\mathbf H^T + \mathbf R)^{-1} \\
\mathbf x = \bar {\mathbf x} + \mathbf K \mathbf y \\
\mathbf P = (\mathbf I - \mathbf K \mathbf H) \bar {\mathbf P}
\end{matrix}
\quad
\right\}& \text{Measurement update}.
\end{aligned}
\end{equation}

EKF is a recursive algorithm working in two steps e.g. prediction and measurement. At the measurement state, using $\hat{X}_k$ and $P_k$ obtained from prediction state, the Kalman Gain $K$ is calculated representing the trustable value of state model and measurement variable. We set measurement noise $\mathbf R$ and process noise $\mathbf Q$ as:

The EKF works recursively in two stages: prediction and measurement. During the measurement phase, the Kalman Gain $K$ is computed using \( \hat{X}_k \) and \( P_k \) obtained from the prediction stage, which signifies the reliability of both the state model and the measurement variable. We define the measurement noise \( \mathbf{R} \) and process noise \( \mathbf{Q} \) as follows:

\begin{equation}
\begin{aligned}
\mathbf R = \begin{bmatrix}1, 0.5, 0.5, 5\end{bmatrix}^T\\
\mathbf Q = \begin{bmatrix} 
1 & 0 & 0 \\ 
0 & 1 & 0 \\
0 & 0 & 1
\end{bmatrix}\dfrac{\sigma_aT^4}{3}
\end{aligned}
\end{equation}

\noindent where $\sigma_a = 1$, and set initial value of $\mathbf P$ as identity matrix. If $R$ tends towards zero, it indicates higher reliability in the measurement variable compared to the state model. Conversely, if $P_k$ tends towards zero, the opposite holds true. Subsequently, the measurements undergo updating based on the $K$ value. The algorithm proceeds through the following steps in the solution.



\begin{algorithm}
	\caption{USV position estimation.}
	\begin{algorithmic}[1]
		\REQUIRE Valid $\mathbf{z} = [r, \alpha, \varepsilon, h]^T$
		\ENSURE New $\mathbf{x}$
		\IF{First frame of $\mathbf{z}$}
		\STATE Calculate initial value of $\mathbf{x}$ with geometric method
		\ELSE
		\STATE Update the gap time $T$, and update $\mathbf{Q}$ correspondingly
		\STATE Do the prediction with $\mathbf{x}$ and $\mathbf{P}$
		\STATE Do the measurement update with $\bar{\mathbf{x}}, \mathbf{z}, \bar{\mathbf{P}}$
		\STATE Output the new $\mathbf{x}$
		\ENDIF
	\end{algorithmic}
\end{algorithm}

\section{Results and Discussion}\label{sec:expsetup}

\subsection{Experimental setup}\label{sec:res}
\begin{figure}[t]
\centering
\includegraphics[width=1\linewidth]{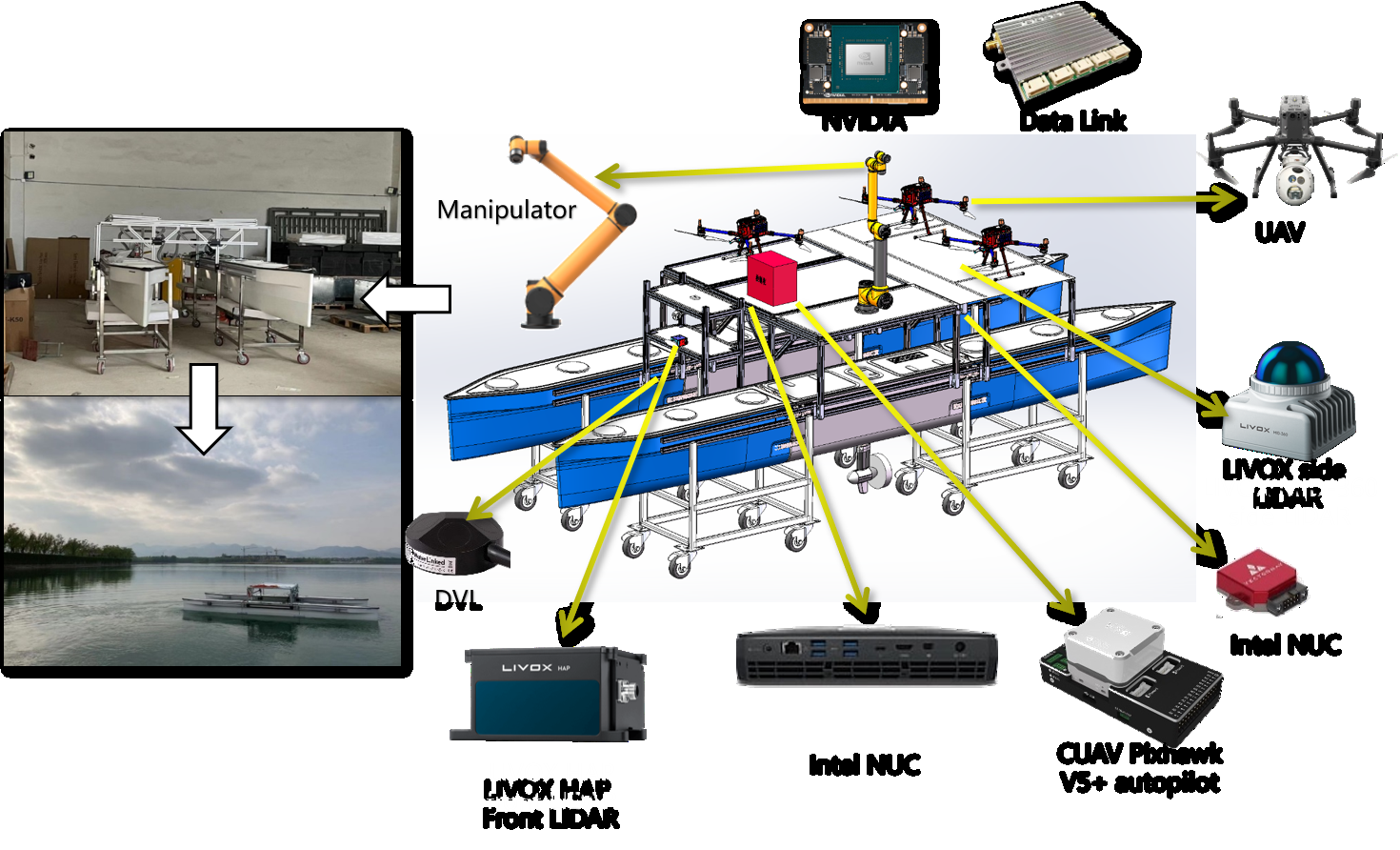}
\caption{The USV system's components. Different sensors such as camera, LiDAR, onboard computer, are installed on the USV to facilitate the autonomous marine operation for searching and intervention tasks.
}
\label{fig:usvhw}
\end{figure}

In our experiment, we employ a USV equipped with different sensors and navigation systems. The USV, developed by Spin Italia S.r.l, is specifically designed for the ``Mohamed Bin Zayed International Robotics Challenge 2024''. It has dual thrusters, a DVL, LiDAR, and an onboard camera. Additionally, our experimental setup incorporates a customized UAV equipped with a Pod camera boasting a field of view ranging from 2.3 to 63.7 degrees, with a distance range of approximately 4 $km^2$. The communication interface is facilitated through Nvidia NX on the USV, which is equipped with 16 GB memory and Ubuntu (20.04) with ROS (Neotic) installed. This setup enables seamless data transmission between the USV, UAV, and our control station, enabling real-time monitoring and control. With its adaptability and reliability, the USV stands as an invaluable asset in our pursuit of advancing marine research and exploration. Fig.~\ref{fig:usvhw} shows the USV system's components and Fig.~\ref{fig:usv} shows the USV in operating mode in real marine environment. 

\subsection{Results}

\begin{figure}[t]
\centering
\includegraphics[width=1\linewidth]{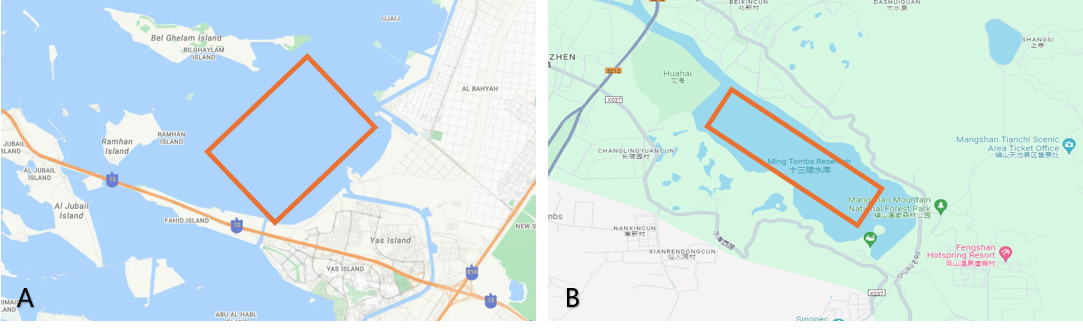}
\caption{Test area. (A) Yas Island, Abu Dhabi, UAV, (B) Shisanling Lake, Changping, Beijing.
}
\label{fig:testarea}
\end{figure}

In this section, we present the experimental results of the proposed vision-based positioning technique. These experiments were carried out at two distinct sites: 1) Shisanling Lake, Changping, Beijing, and 2) Yas Island, Abu Dhabi, UAE (see Fig.~\ref{fig:testarea}. The primary aim of these experiments was to investigate the effectiveness of the proposed method in estimating the position of the USV when compared to a GPS. For experiments, we discuss how it was conducted and the obtained results. Lastly, we discuss some remarks on the system's performance.

In experiment 1, we evaluate the efficacy of the proposed method within a lake environment. Initially, we execute a series of experiments employing an open-loop control approach, wherein the USV is manually navigated along predefined $x$ and $y$ directions for a certain period. The position of the USV was recorded via the proposed scheme and compared with the onboard GPS. The proposed scheme produces relatively similar results to those of GPS. In the subsequent scenario, the USV is directed to navigate toward predetermined $x$ and $y$ locations employing a closed-loop control algorithm,  where the vehicle is operating at a certain surge speed for some time. We plot the USV position and yaw angle for all scenarios and compare them with the GPS measurements.

Experiment 2 extends the scope to assess how the proposed approach performs under extreme weather conditions in a real marine environment. These tests were conducted at sea with ocean disturbances, wind, and light reflections. The test area spanned approximately 4 \time 4 $km^2$ at Yas Island, Abu Dhabi, UAE. We explore how the UAV detects the USV when located too far from the coastal area where no GPS signals are available. Subsequently, we show how our proposed method eliminates drift and provides a smooth position measurement of the USV during the tracking process.


 \begin{figure}[t]
\centering
\includegraphics[width=1\linewidth]{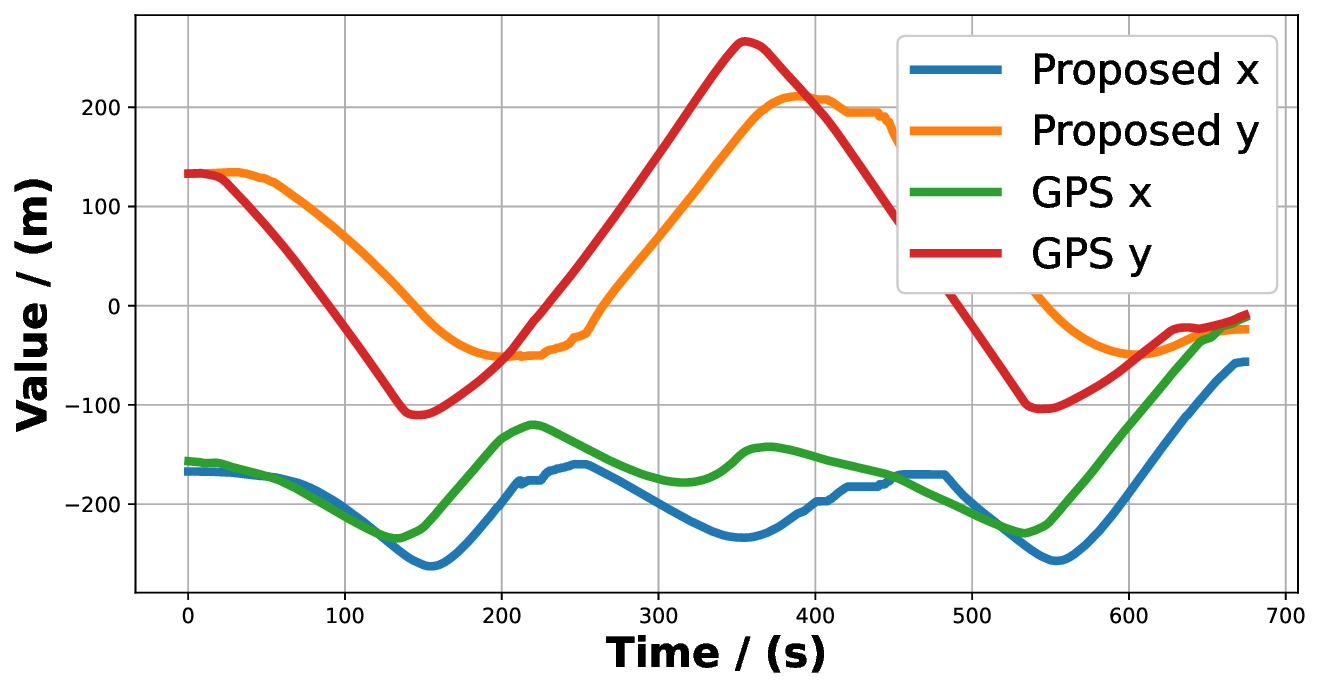}
\caption{The estimated and real $x$ and $y$ positions of the USV using the proposed method. These results are from scenario 1 of experiment 1 carried out at BIT, China.
}
\label{fig:pos1}
\end{figure}

 \begin{figure}[t]
\centering
\includegraphics[width=1\linewidth]{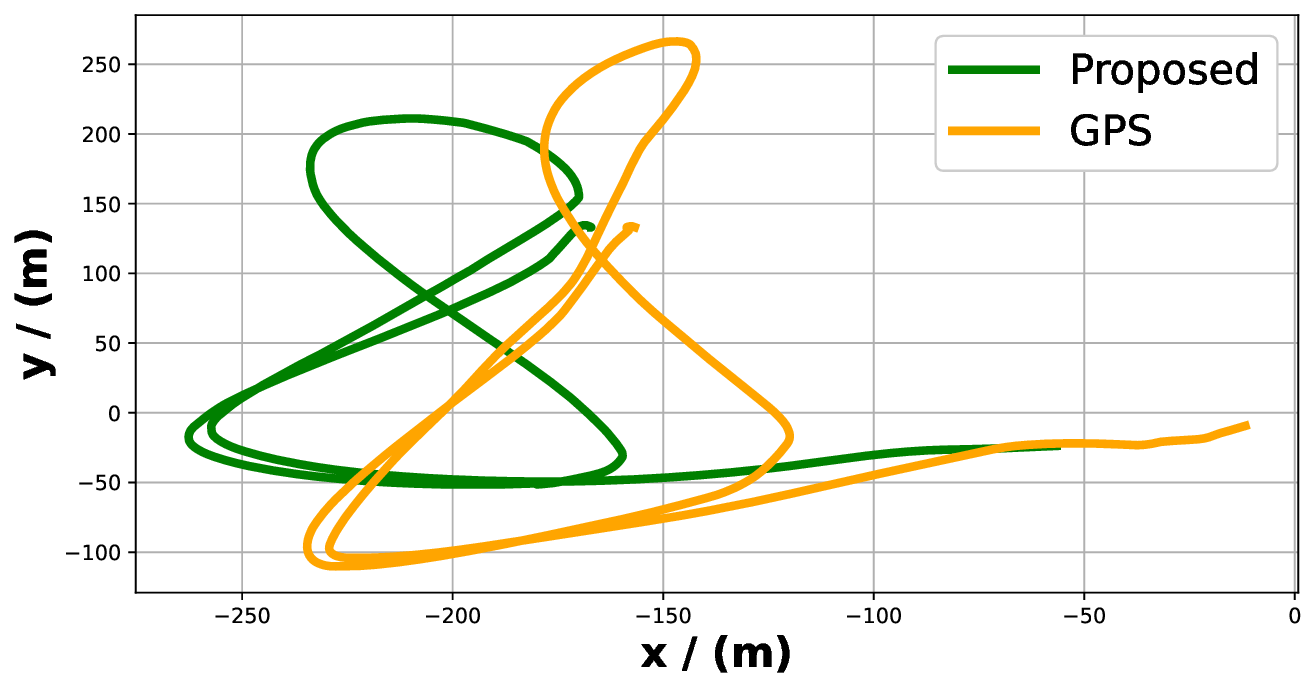}
\caption{The $x$ versus $y$ positions of the USV estimated by the proposed method were compared with the actual positions. These results are from scenario 1 of experiment 1 conducted at BIT, China. The green line represents the results of the proposed method, while the orange line depicts the GPS recorded measurements.
}
\label{fig:pos1xy}
\end{figure}


 \begin{figure}[t]
\centering
\includegraphics[width=1\linewidth]{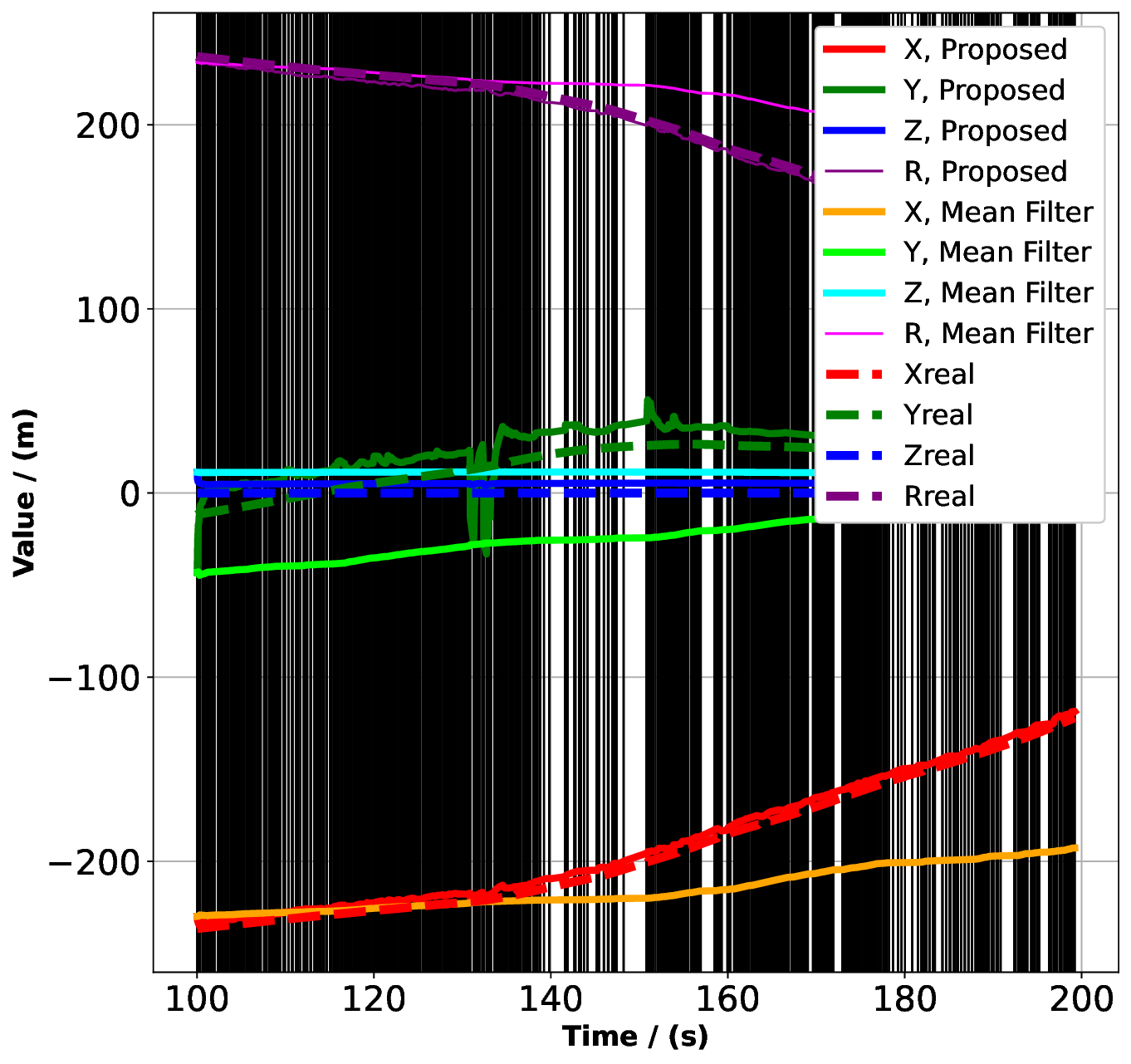}
\caption{The positions ($x$, $y$, and $R$) of the USV obtained using the proposed method, mean filter method, and GPS measurements are compared based on real data from scenario 2 of experiment 1 conducted at BIT, China.
}
\label{fig:pos2}
\end{figure}

 \begin{figure}[t]
\centering
\includegraphics[width=1\linewidth]{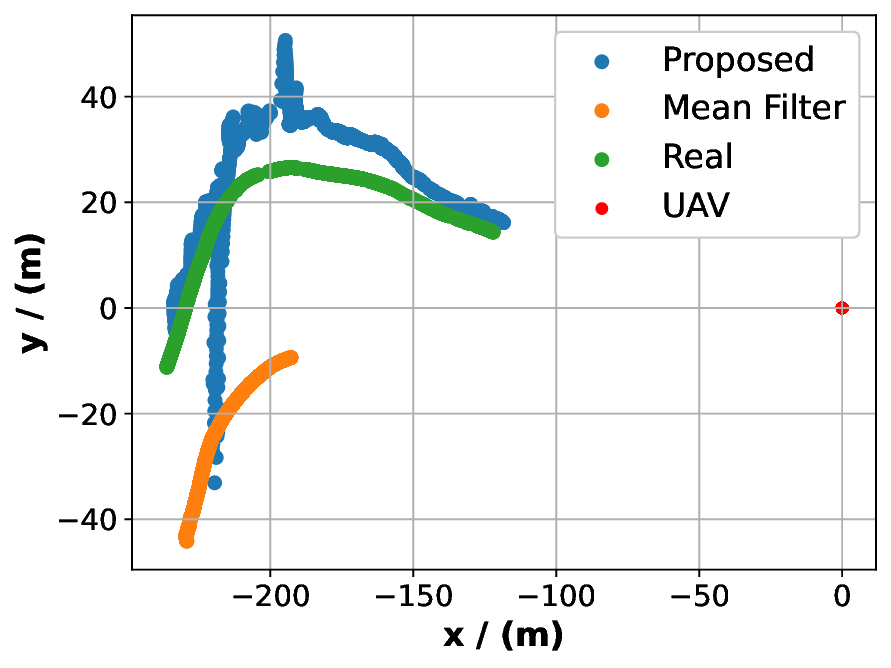}
\caption{The x versus y positions of the USV were compared using results obtained from the proposed method, mean filter method, and real data from scenario 2 of experiment 1 conducted at BIT, China.
}
\label{fig:pos2xy}
\end{figure}

 \begin{figure}[t]
\centering
\includegraphics[width=1\linewidth]{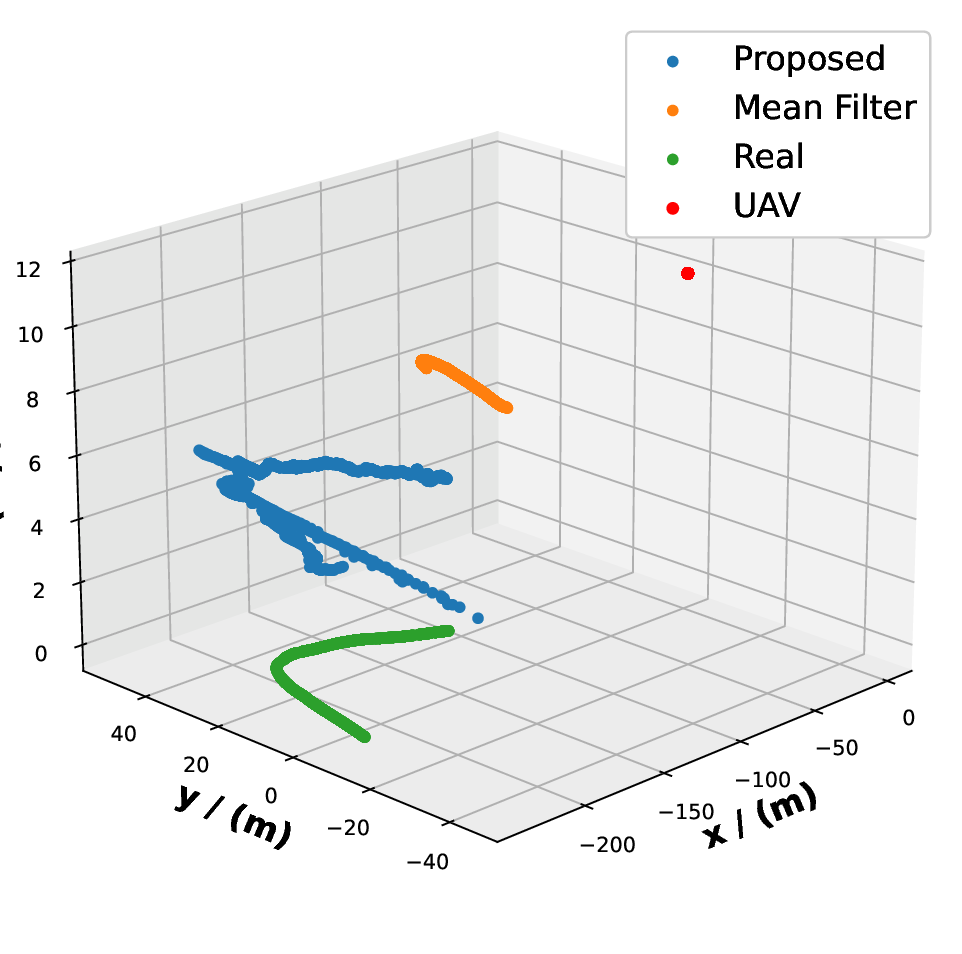}
\caption{The results from scenario 2 in experiment 1 showing $x,y$ and $x$ through the proposed method, mean filter method, and the GPS recorded measurement. 
}
\label{fig:pos2xy3d}
\end{figure}

 \begin{figure}[t]
\centering
\includegraphics[width=1\linewidth]{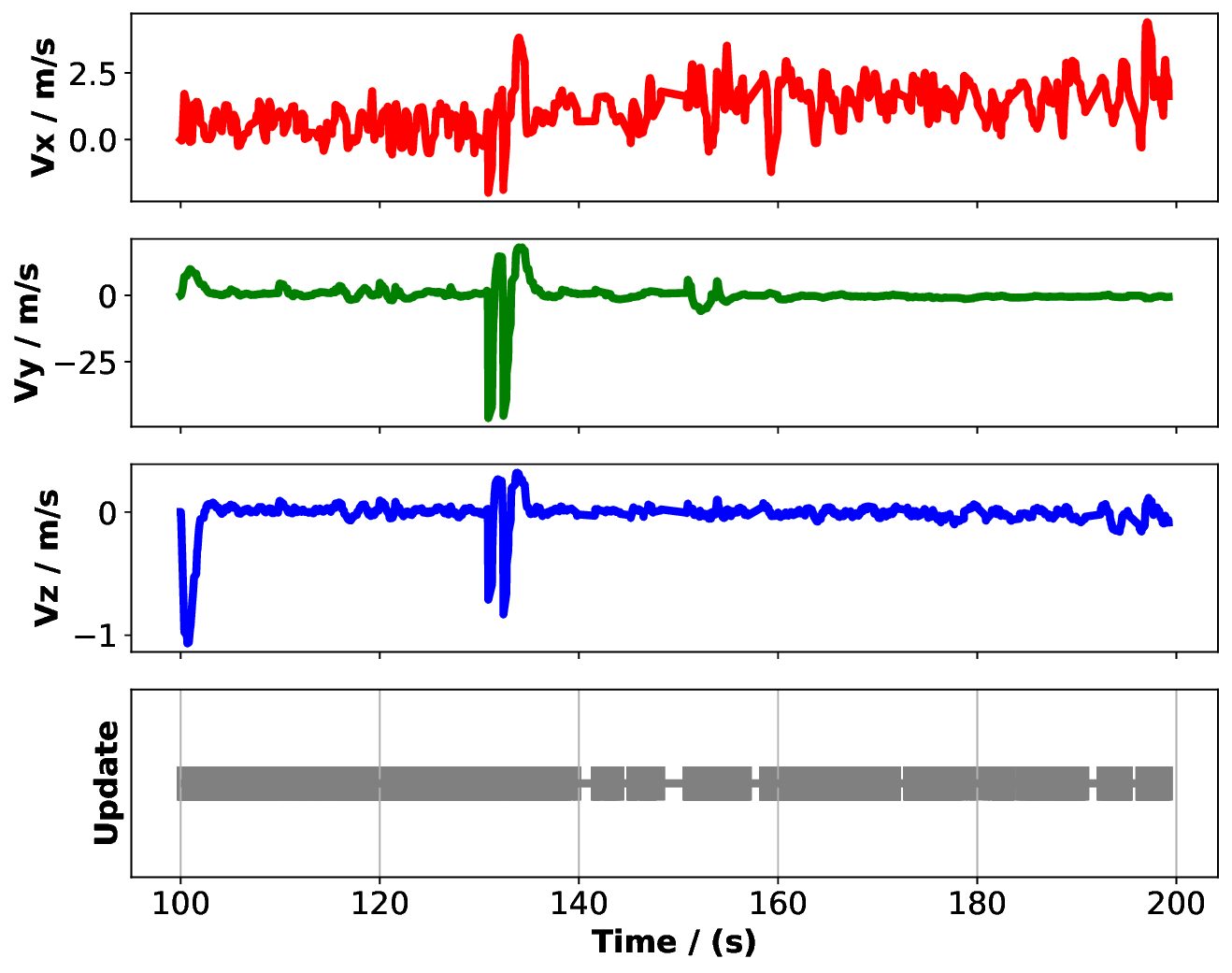}
\caption{The velocity profile of the USV utilized in scenario 1 of experiment 2 incorporates all velocities—red (Vx) and green (Vy)—to guide the USV towards the target position. The results indicate that despite rapid changes in the velocity profile, the proposed method generates smooth positioning curves.
}
\label{fig:pos2vel}
\end{figure}

 \begin{figure}[t]
\centering
\includegraphics[width=1\linewidth]{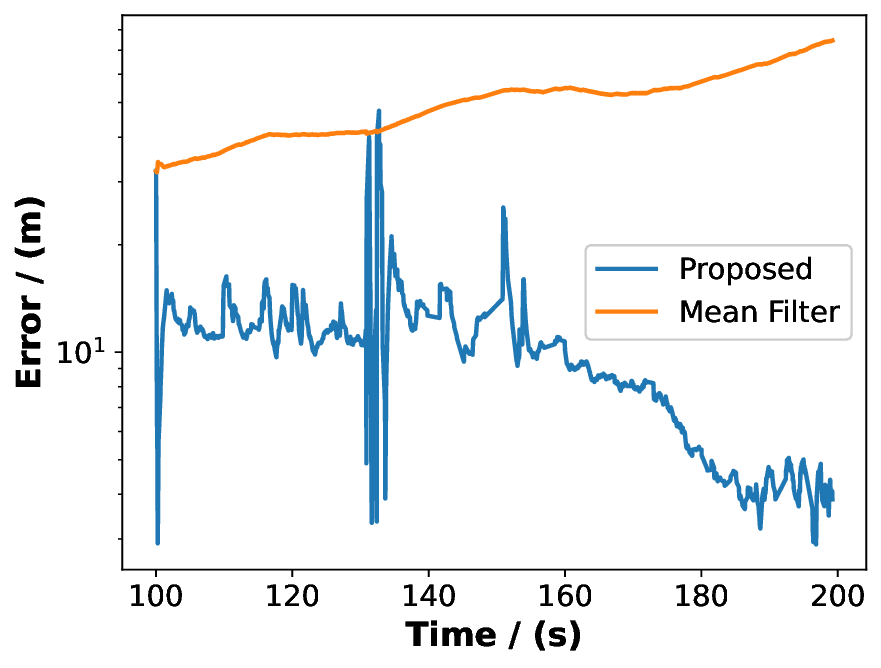}
\caption{Comparison of Error between the proposed method and the mean filter method. The results show the error analysis between the mean filter (orange line) and proposed EKF (blue line) for scenario 2 in experiment 1. The proposed method demonstrates consistently lower error rates across the operation, highlighting its better performance compared to the mean filter method.
}
\label{fig:pos2error}
\end{figure}


\begin{figure}[t]
\centering
\includegraphics[width=1\linewidth]{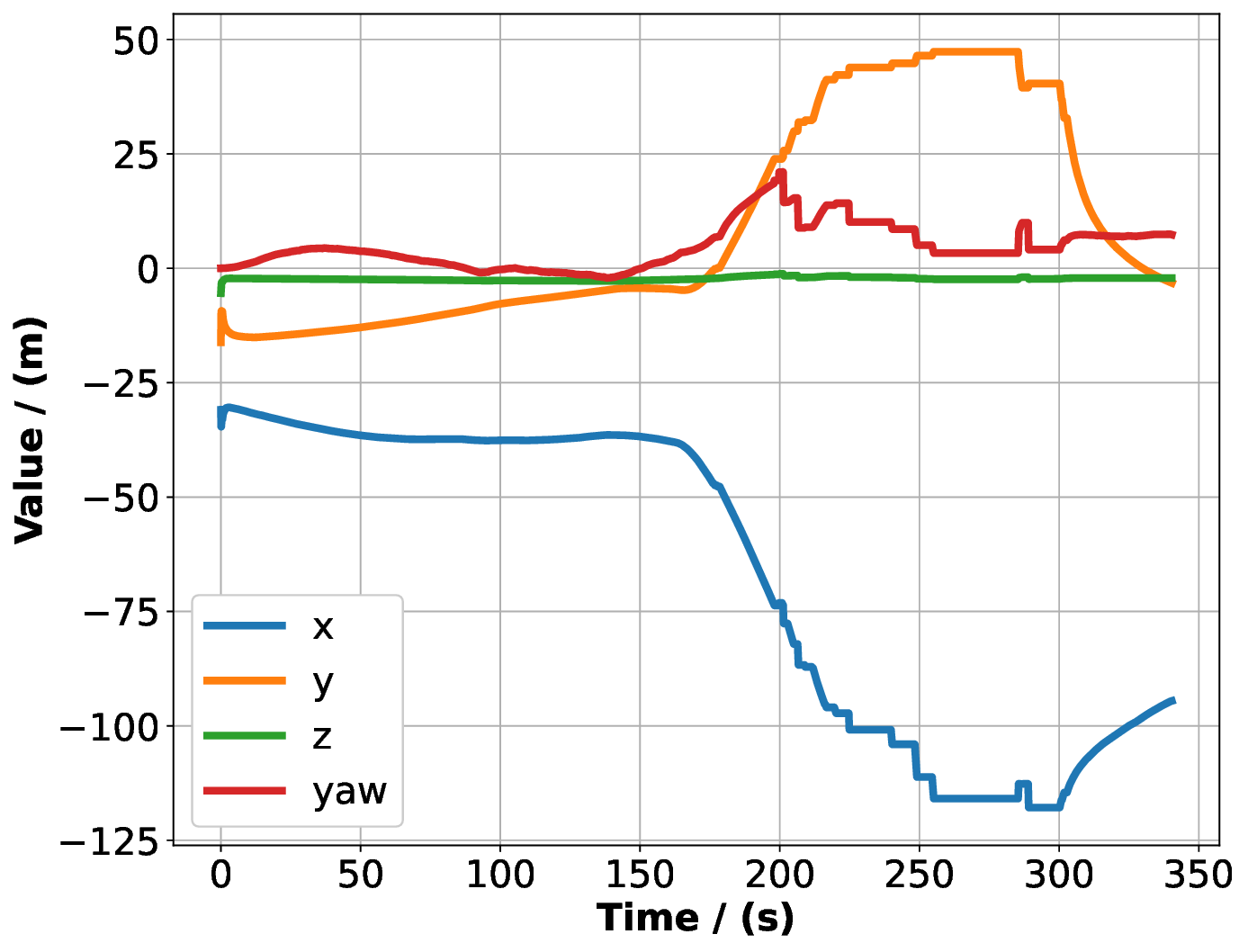}
\caption{The results of run 1 in the MBZIRC-2024 competition illustrate the USV's ($x$, $y$, $z$, and yaw) positions during operation. The results demonstrate the successful generation of accurate position data at long ranges of up to 500 meters using the proposed method.
}
\label{fig:pos3}
\end{figure}

\begin{figure}[t]
\centering
\includegraphics[width=1\linewidth]{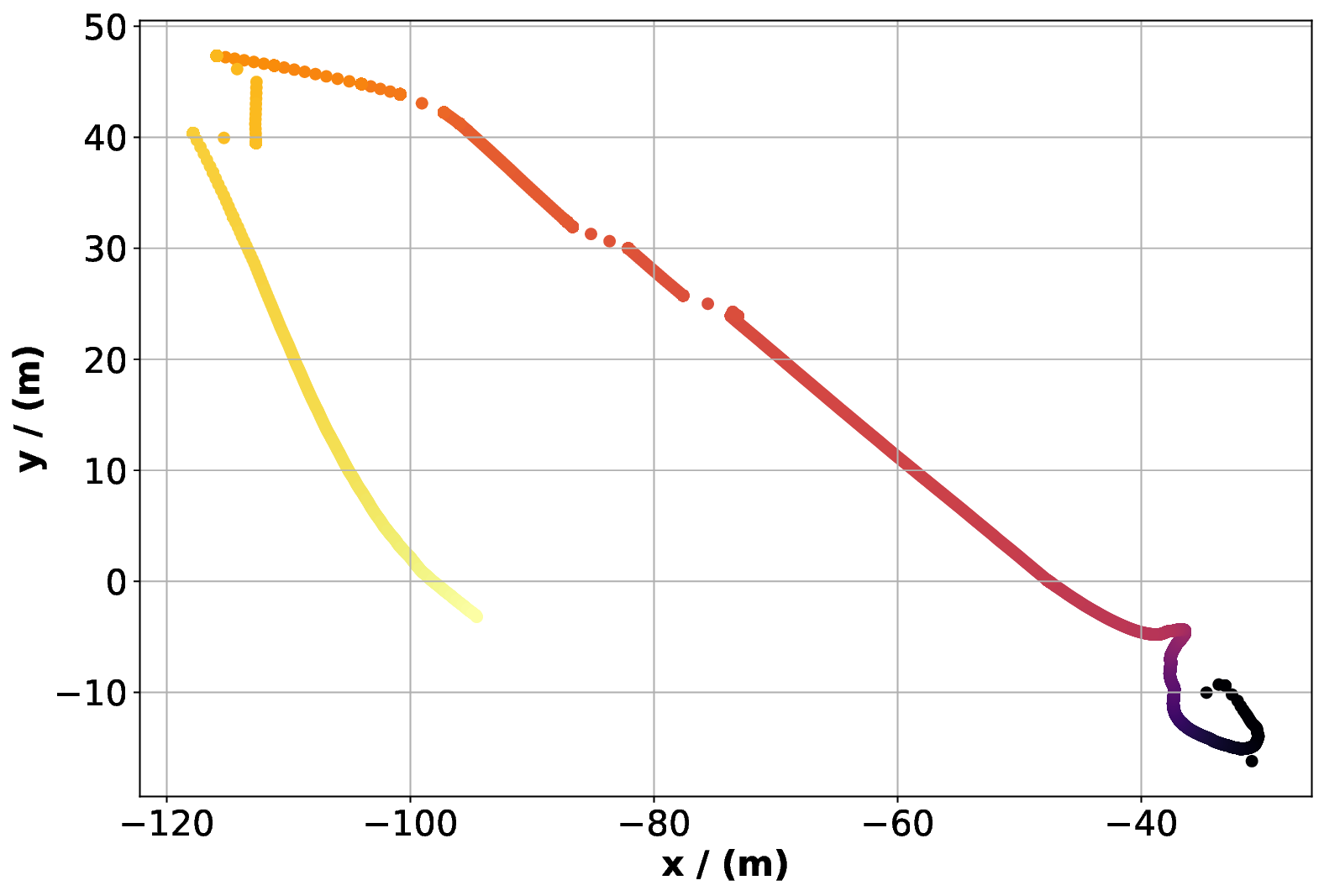}
\caption{USV x versus y position of the USV trajectory during run 1 in the MBZIRC-2024 competition. The results visualize the travelled path in the north and east direction during the experiment.
}
\label{fig:pos3xy}
\end{figure}

\begin{figure}[t]
\centering
\includegraphics[width=1\linewidth]{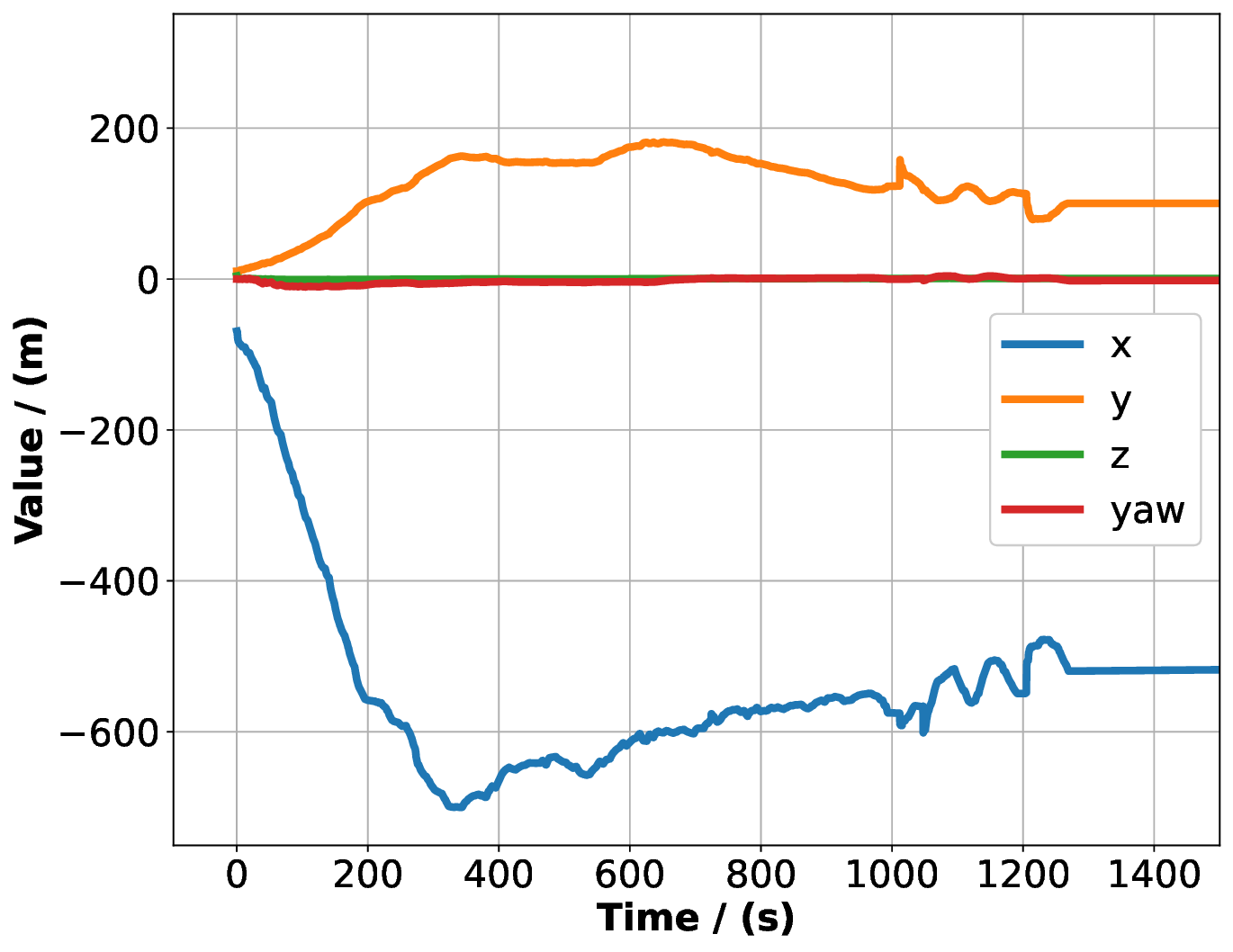}
\caption{The results form run 2 in MBZIRC-2024 competition showing USV $x$,$y$,$x$, and $yaw$ positions. The results demonstrate the USV trajectory travelled during the competition showcasing the long-range capability of the proposed method for USV localization.
}
\label{fig:pos4}
\end{figure}

\begin{figure}[t]
\centering
\includegraphics[width=1\linewidth]{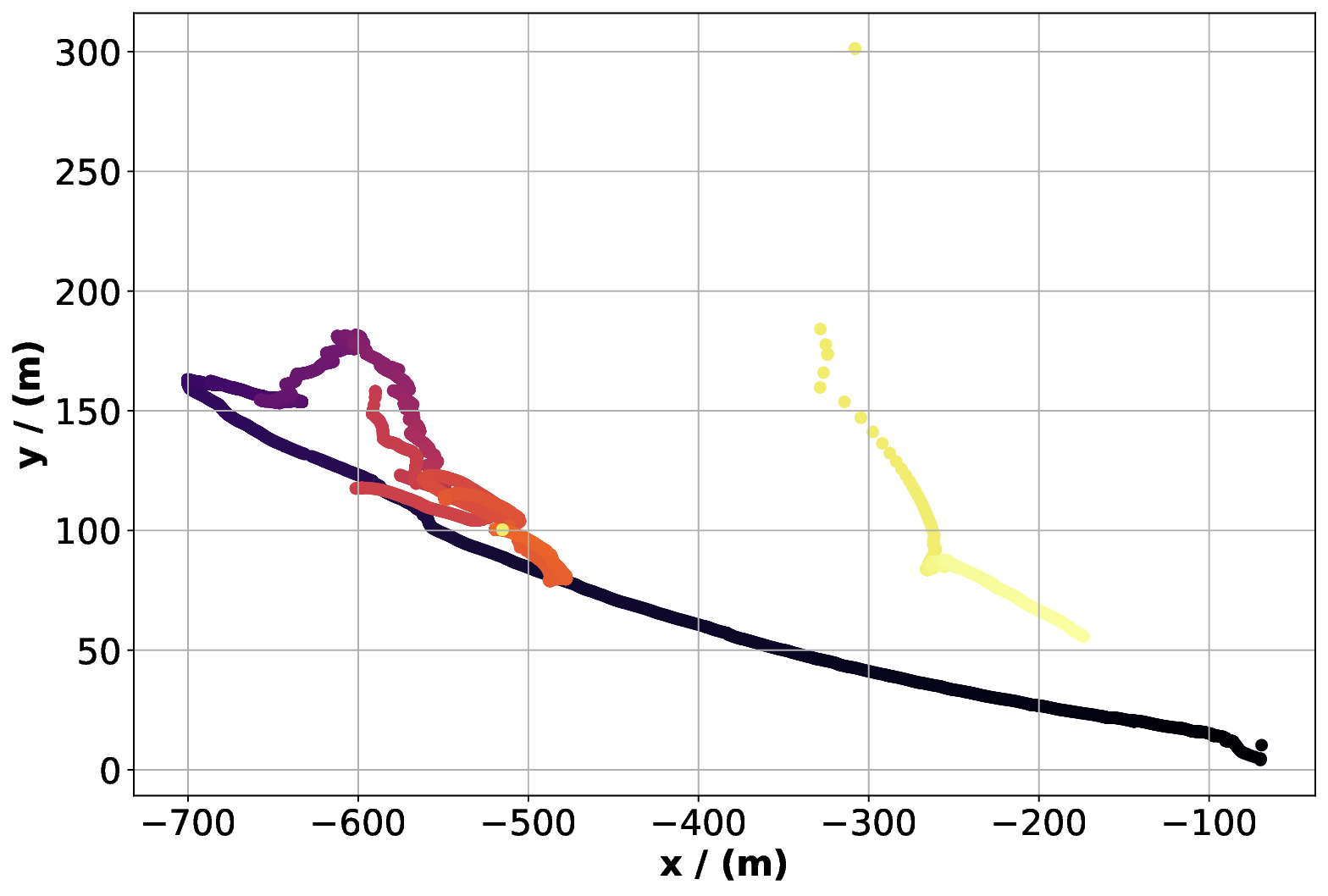}
\caption{The results of the USV x versus y position illustrate the trajectory travelled during run 2 of MBZIRC-2024. It depicts the USVI's navigation starting from the origin, moving north for up to 700 meters, and then east for up to 200 meters.
}
\label{fig:pos4xy}
\end{figure}


The results from scenario 1 in experiment 1 are shown in Figs.~\ref{fig:pos1} and \ref{fig:pos1xy}. Fig.~\ref{fig:pos1} shows the trajectory of the USV as it navigates in the lake. The x-axis represents time, while the y-axis indicates the position of the USV in meters. The plot illustrates the USV's movement over a specified duration, showcasing its traveled distance in $x$ and $y$ direction. The results demonstrated that the vehicle position extracted through the proposed vision-based method is similar to the USV's onboard GPS.
Similarly, Fig.~\ref{fig:pos1xy} shows the $x$ and $y$ position of the USV compared with the GPS measurement. The orange line shows the GPS data, and the green line shows the proposed EKF measurements. The results show that the proposed scheme has demonstrated better performance for the USV localization in the lake setup.

The results from scenario 2 in experiment 1 are shown in Figs.~\ref{fig:pos2}-\ref{fig:pos2error}. The USV estimated position in the ENU frame is plotted in Fig.~\ref{fig:pos2}. Here, we plotted measurements of three sources: EKF, Mean Filter, and GPS. The GPS measurements of $X, Y, Z$, and $R$ are considered ground truth for comparison purposes. The grey vertical lines means how many times we got valid measurements. In addition to EKF, we also used the Mean Filter technique for position estimation. The results demonstrated that the proposed EKF method successfully tracked and followed the ground truth. However, the results from the Mean filter showed deviation in the USV positions compared with GPS measurement. The results indicate that the proposed method can supplement the USV localization in a GNSS-denied environment. 

Figs.~\ref{fig:pos2xy} and \ref{fig:pos2xy3d} show the estimated relative USV ENU position to UAV with a) proposed method, b) mean filter estimated, and c) real GPS data as ground truth. The red point at the origin shows the UAV position and is considered a starting point. The trajectory of GPS data indicates that the USV first went towards the northeast and then turned towards the southeast. The output trajectory of the proposed method converged towards the actual trajectory, while the mean filter didn't eliminate estimation error. From the results, it is noticed that the trajectory obtained by the proposed method closely aligns with the actual trajectory, highlighting the method's capability to estimate the relative position of the USV accurately relative to the UAV. The results demonstrated that the proposed method could provide reliable guidance and tracking information for the USV, facilitating tasks such as surveillance, monitoring, or coordination in marine operations.

Fig.~\ref{fig:pos2vel} shows the estimated velocity of the USV throughout the experiment. The x-axis depicts the time in seconds, while the y-axis represents the vehicle's velocity in meters per second (m/s). The plot provides insights into the dynamic behavior of the USV as it maneuvers in the lake during the experiment. Although there is some fluctuation in the surge and sway speed of the vehicle, the proposed method can still eliminate the error and produce accurate localization results comparable with the ground truth measurements. 


Fig.~\ref{fig:pos2error} shows the absolute trajectory error obtained by the proposed method and the mean filter method. The results reveal a notable difference in error between the two approaches, with the proposed method exhibiting significantly lower error than the mean filter method. Specifically, the proposed method demonstrates an average localization error of 2.76 meters and a maximum localization error of 4.76 meters. In contrast, the mean filter method fails to mitigate drift, resulting in a persistent deviation from the ground truth trajectory. A comparison with the ground truth highlights the enhanced localization accuracy achieved by the proposed method. These findings demonstrate the efficacy of the proposed approach in meeting localization requirements and providing robust and precise USV positioning in GNSS-denied marine environments.

In Experiment 2, we implemented the proposed method in a real marine environment at Yas Island, Abu Dhabi, UAE, where GPS signals were restricted. The results of Experiment 2 are depicted in Figs.~\ref{fig:pos3}-\ref{fig:pos4xy}. These results are extracted from two different runs conducted as part of the MBZIRC-2024 Challenge, which focused on developing and evaluating solutions requiring coordinated efforts between UAVs and USVs to execute complex navigation and manipulation tasks in GNSS-denied marine environments. The testing area spanned 4x4 $km^2$. The challenge entailed a series of tasks: first, locating a target vessel within the marine environment, followed by guiding a USV to approach and dock at this vessel. Subsequently, UAVs stationed on the USV flew out to inspect small suspicious boxes aboard the target vessel. Collaboratively, the UAVs retrieved these boxes and transported them back to the USV. Finally, the last task involved using the manipulator attached to one side of the USV to grasp and transfer a large box from the target vessel onto the USV.

The proposed vision-based and UAV-assisted localization method was utilized during the operation. From a constant height of around 7.5m, the UAV was allowed to search and scan the marine area for the target vessel. Once the vessel's position was identified, the UAV switched to the USV's localization mode. Then, as long as the USV was approaching the target location, the UAV continuously and synchronously performed the localization tasks of the USV. Fig.~\ref{fig:pos3} shows the position $x, y, z$, and angle $yaw$ in ENU frame during the run 1 from experiment 2. The origin of the USV was considered the current docking position. From the results, it can be noted that the proposed method produces smooth position curves along $x, y$ directions. It can be more clearly observed in Fig.~\ref{fig:pos3xy} where $x,y$ position are plotted.
Similarly, Figs.~\ref{fig:pos4} and \ref{fig:pos4xy} show the results obtained during run 2 from experiment 2. In this case, the USV traveled a very long distance, up to 1$km^2$, in a forward direction toward the target vessel. The proposed method produces smooth and robust position information of the USV even in the presence of ocean waves, wind, and light reflection. During the operation, the vehicle accelerates significantly to reach the target position. As a result, the position curves show some rapid drift for a short period, as shown in Fig. \ref{fig:pos4} after the 1800s. However, the proposed method reduces drift and produces a stable position measurement.

\begin{table}[t]
    \centering
    \caption{Performance comparison of the proposed approach with Mean Filter and No Filter during Experiment 1 and Experiment 2. The Error for $X,Y$, and $\text{2D}$ are recorded with respect to time/s ($10,50,100)$.
    }
    \begin{tabular}{cccccc}
    \hline
    
    \hline

    Test & Strategy & Time (s) & X & Y  & 2D\\\hline
    
    \hline
    Exp 1 & Proposed & 10 & -0.013    & -0.012     & 0.018  \\
     &  & 50 & -0.003      & -0.012      & 0.012  \\
     &  & 100 &  0.014       & 0.023       & 0.028 \\\cline{2-6}

     & Mean Filter & 10 &  0.001     & -55.485     &55.485  \\
     &  & 50 & -16.496    & -54.779   & 57.209\\
     &  & 100 & -18.682    & -66.389    & 72.307  \\\cline{2-6}

      & No Filter & 10 & -0.178    & -60.077    & 60.077 \\
     &  & 50 & -16.381    & -44.628 & 47.540 \\
     &  & 100 & -18.541    & -69.890  & 72.307  \\\hline

     Exp 2 & Proposed & 10 & 3.84     & 15.84     & 16.30 \\
     &  & 50 & 4.72       & 13.25     & 14.07 \\
     &  & 100 &  3.66     & 1.79      & 4.07 \\\cline{2-6}

     & Mean Filter & 10 & 3.48    & -36.76   & 36.92 \\
     &  & 50 & -19.80    & -50.25   & 54.01 \\
     &  & 100 &-70.75     & -23.77     & 74.64 \\\cline{2-6}

      & No Filter & 10 &  5.40      & -32.35    & 32.80  \\
     &  & 50 & 0.83      & -30.27    & 30.29  \\
     &  & 100 & 1.48   & -10.59 & 10.69  \\

    \\ \hline

    \hline
    \end{tabular}
    \label{tab:scores-datasets}
\end{table}

The performance comparison of the proposed approach with the Mean Filter and No Filter during Experiment 1 and Experiment 2 showed notable differences in error reduction over time. Table~\ref{tab:scores-datasets} shows the results for both experiments, errors in the $X$, $Y$, and $\text{2D}$ measurements were recorded at time intervals of $10$, $50$, and $100$ seconds. In Experiment 1, the proposed approach consistently outperformed both the Mean Filter and the No Filter configurations across all time intervals. Experiment 2 demonstrated a similar pattern, with the proposed approach again showing better performance. However, the overall error magnitudes were slightly higher compared to Experiment 1, possibly due to more complex environmental conditions or different experimental parameters. The proposed method maintained its advantage over the Mean Filter and No Filter approaches, with the lowest errors recorded at each time step. The Mean Filter showed some ability to mitigate errors compared to No Filter, but it was less effective in consistently reducing the $\text{2D}$ error over time. The proposed approach demonstrated robust error minimization capabilities across both experiments, highlighting its effectiveness in improving accuracy over time compared to traditional filtering methods.










\subsection{Discussion}
In this work, we presented a UAV-assisted vision-based localization method for the USV in the marine environment in a GNSS-denied environment. We conducted experiments to test and validate the method's performance in different test locations under various conditions. The results obtained from experiment 1 and experiment 2 verify the method's applicability for USV localization in the real-time marine environment. 

In Experiment 1, we showed the performance of our localization method by comparing it with ground truth GPS measurements, illustrated in Figs.~\ref{fig:pos1}-\ref{fig:pos2error}. Multiple test runs revealed that our method accurately tracked the GPS-recorded positions. To further validate our approach, we compared it with a mean filter and observed that it failed to accurately follow the ground truth measurements during operation. Additionally, we demonstrated that the positioning measurements obtained from our method exhibit smoothness and robustness under the velocity profile used for USV maneuvering. Our process also exhibited lower position errors compared to the mean filter approach.

In Experiment 2, we further extend the scope and showcase the performance of our localization method for inspection and intervention tasks using a heterogeneous collaborative system in a GNSS-denied environment. The results obtained (refer to Figs.~\ref{fig:pos3}-\ref{fig:pos4xy}) demonstrated the ability of our method to accurately estimate the USV position in the marine environment using visual information. Particularly noteworthy was the observation that our method enabled drift-free USV localization tasks—using a UAV equipped with a camera significantly extended operational time and range. The UAV's camera could scan marine areas up to 4$km^2$, facilitating localization of the USV over long distances. Since experiment 2 was conducted during the MBZIRC-2024 competition in a GNSS-denied environment, we could not directly benchmark the results of the proposed method with the GPS measurement. Nonetheless, the experimental results highlighted that our method effectively provided the necessary position information for USV navigation to achieve target positions during the experiments.


The maximum distance to locate any target using the proposed algorithm largely depends on the effectiveness of the vision detection algorithm and the capabilities of the photoelectric pod or camera. This is because the image serves as the sole source of directional information toward the target. For instance, the data link range can provide consistent distance information when attempting to locate the USV with a cooperative target using a small UAV. However, suppose the target appears too small in the image, even when the camera is zoomed to its maximum horizontal field of view. In that case, no directional information about the target can be obtained, leading to algorithm failure.

The algorithm operates effectively when the camera has a direct line of sight to the target, which can be compromised if obstacles block the view. During MBZIRC 2024, the small UAV maintained a hover height of 7.5 meters, and the targets were located at distances ranging from approximately 50 meters to 2000 meters. Additionally, given that there were a maximum of 8 vessels (including target vessels and USV) in the competition sea area, the likelihood of these vessels aligning in the same direction relative to the UAV was extremely low. Therefore, when considering the application of the proposed algorithm in experiments, careful attention should be paid to the height-to-distance ratio and the number of vessels on the sea.

\section{Conclusion}\label{sec:conclusion}
In this paper, we showed how a vision-based approach assisted by an UAV can perform localization tasks and improve the USV navigation in a large marine environment where GNSS signals are unavailable. Our method involves a heterogeneous system comprising of an UAV for scanning and localizing the USV, employing vision-based techniques to determine the USV's relative position with respect to the UAV. The effectiveness of our approach was successfully demonstrated in both lake and real marine environments. The results indicate that our proposed method provides accurate and smooth positioning measurements of the USV while navigating in marine environments, even in the presence of wind and light reflections. Notably, our method demonstrated the capability to localize the USV over a large marine area spanning approximately 4$km^2$ (as shown in the MBZIRC-2024 challenge). However, the proposed method is not trouble-free. For instance, localization relies solely on visual data captured by the UAV's camera. Localization may fail if the USV is not clearly visible or appears too small in the images. These limitations could be addressed in the future by employing sensor fusion techniques for USV detection. Furthermore, the presence of other objects in the captured images can lead to localization errors if the model incorrectly identifies them as the USV. In future work, we aim to address these issues and conduct closed-loop experiments to demonstrate the method's applicability in more complex marine environments.

\section*{Acknowledgement}
\noindent This work is supported by the Khalifa University under Award No. RC1-2018-KUCARS-8474000136, CIRA-2021-085, MBZIRC-8434000194, KU-BIT-Joint-Lab-8434000534, and by Beijing Institute of Technology under National Key Research and Development Program under Grant No. 2022YFE0204400.

\bibliographystyle{IEEEtran}
\bibliography{ref}

\end{document}